\documentclass[10pt,twocolumn,letterpaper]{article}

\usepackage{iccv}
\usepackage{times}
\usepackage{epsfig}
\usepackage{graphicx}
\usepackage{amsmath}
\usepackage{amssymb}
\usepackage[svgnames]{xcolor}
\usepackage{xcolor}
\usepackage{soul}
\usepackage{sectsty}

\usepackage{bm}

\usepackage{array,multirow}
\usepackage{float}

\DeclareMathOperator*{\argmin}{arg\,min}

\usepackage[linesnumbered,ruled,vlined]{algorithm2e}
\usepackage{algpseudocode}
\usepackage{amsmath}

\usepackage[breaklinks=true,bookmarks=false]{hyperref}

\iccvfinalcopy 

\def\iccvPaperID{****} 

\ificcvfinal\fi

\begin{document}

\title{\vspace{-.06in}AGKD-BML: Defense Against Adversarial Attack by Attention Guided Knowledge Distillation and Bi-directional Metric Learning}


\author{Hong Wang$^{1}$\thanks{This work was done during the research assistantship at BNL.}, Yuefan Deng$^{1}$, Shinjae Yoo$^{2}$, Haibin Ling$^{1}$, Yuewei Lin$^{2}$\thanks{Corresponding author.}\\
$^{1}$Stony Brook University, Stony Brook, NY, USA\\$^{2}$Brookhaven National Laboratory, Upton, NY, USA\\
{\tt\small \{hong.wang.2,yuefan.deng,haibin.ling\}@stonybrook.edu,\,\,\,\{sjyoo,ywlin\}@bnl.gov}
}

\maketitle
\ificcvfinal\thispagestyle{empty}\fi

\begin{abstract}
While deep neural networks have shown impressive performance in many tasks, they are fragile to  carefully designed adversarial attacks. We propose a novel adversarial training-based model by \emph{Attention Guided Knowledge Distillation} and \emph{Bi-directional Metric Learning} (AGKD-BML). The attention knowledge is obtained from a weight-fixed model trained on a clean dataset, referred to as a teacher model, and transferred to a model that is under training on adversarial examples (AEs), referred to as a student model. In this way, the student model is able to focus on the correct region, as well as correcting the intermediate features corrupted by AEs to eventually improve the model accuracy. Moreover, to efficiently regularize the representation in feature space, we propose a bidirectional metric learning. Specifically, given a clean image, it is first attacked to its most confusing class to get the forward AE. A clean image in the most confusing class is then randomly picked and attacked back to the original class to get the backward AE. A triplet loss is then used to shorten the representation distance between original image and its AE, while enlarge that between the forward and backward AEs. 
We conduct extensive adversarial robustness experiments on two widely used datasets 
with different attacks. Our proposed AGKD-BML model consistently outperforms the state-of-the-art approaches. The code of AGKD-BML will be available at: \url{https://github.com/hongw579/AGKD-BML}.


\end{abstract}

\section{Introduction}

\begin{figure}[tbp]
  \centering
  \includegraphics[width=.95\linewidth]{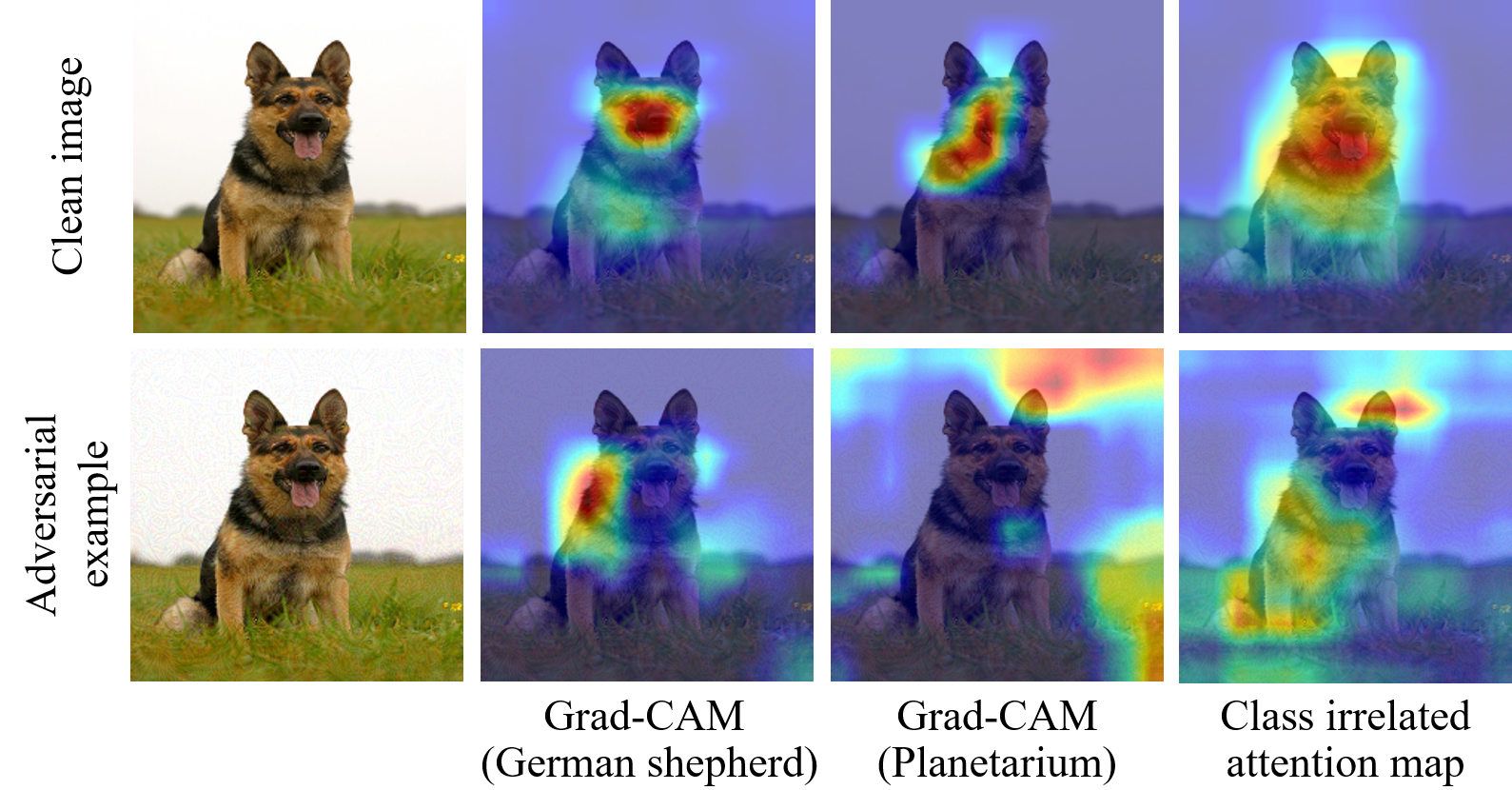}
  \caption{A clean image (``German shepherd”) and its adversarial example (incorrectly classified as ``Planetarium”) are in the first column. The class relevant attention maps (Grad-CAM) of correct and incorrect labels, and the class irrelevant attention maps are shown in the second, third and fourth columns, respectively. It shows that the adversarial perturbations corrupt the attention maps.}
  \label{fig:attention}
\end{figure}

Deep neural networks (DNNs) have achieved great breakthrough on a variety of fields, such as computer vision~\cite{NIPS2012_c399862d}, speech recognition~\cite{hinton2012deep}, and natural language processing~\cite{collobert2008unified}.
However, their vulnerability against the so-called adversarial examples (AEs), which are the data with carefully designed but imperceptible perturbations added, has drawn significant attention~\cite{szegedy2013intriguing}. 
The existing of AEs is a potential threat for the safety and security of DNNs in real-world applications.
Thus, many efforts have been made to defend against adversarial attacks as well as improve the adversarial robustness of the machine learning model. In particular, adversarial training~\cite{goodfellow2014explaining, madry2017towards}-based models are among the most effective and popular defending methods. Adversarial training solves a min-max optimization problem, in which the inner problem is to find the strongest AE within an $\epsilon-$ball by maximizing the loss function, while the outer problem is to minimize the classification loss of the AE. Madry \etal~\cite{madry2017towards} provided a multi-step projected gradient descent (PGD) model, which has become the standard model of the adversarial training. Following PGD, a number of recent works have been proposed to improve adversarial training from different aspects, \eg, \cite{cheng2020cat, ding2020mma, Mao2019, rakin2019bit, shafahi2019adversarial, wang2019convergence, zhang2019defense, zhang2019theoretically}. 

However, the adversarial training-based models still suffer from relatively poor generalization on both clean and adversarial examples. Most of the existing adversarial training based models focus only on the on-training model that utilizes adversarial examples, which may be corrupted, but have not well explored the information from the model trained on clean images. In this work, we aim to improve the model adversarial robustness by distilling the attention knowledge and utilizing bi-directional metric learning. 

The attention mechanism plays a critical role in human visual system and is widely used in a variety of application tasks~\cite{selvaraju2017grad, zhou2016learning}. Unfortunately, one of our observations shows that the perturbations in the adversarial example (AE) will be augmented through the network, and thus significantly corrupts the intermediate features and attention maps. It is shown in the Figure~\ref{fig:attention}, the AE confuses the model by letting it focus on different regions from the clean image. Intuitively, if we can transfer the knowledge of clean images from the teacher model to the student model to 1) obtain right attention information, and 2) correct the intermediate features corrupted by AE, we should be able to improve the model's adversarial robustness. 


With this motivation, we propose an \textit{Attention Guided Knowledge Distillation} (AGKD) module, which applies knowledge distillation (KD)~\cite{hinton2015distilling} to efficiently transfer attention knowledge of the corresponding clean image from the teacher model to the on-training student model. Specifically, the teacher model is pre-trained on the original clean images and will be fixed during training, while the student model is the on-training model. The attention map of a clean image obtained from the teacher model is used to guide the student model to generate the attention map of the corresponding AE against the perturbations.

We further use t-distributed Stochastic Neighbor Embedding (t-SNE) to study the behavior of the AE in the latent feature space (see Figure~\ref{fig:tsne}), and observe that the representations of the AE are usually far away from their original class, similar as shown in~\cite{Mao2019}. While AGKD transfers information of clean image to the student model from the teacher model and thus provides the constraints on the similarity between the AE and its corresponding clean image, there is no constraint of samples from different classes taken into account. Previous works~\cite{li2019improving, Mao2019, Zhong2019} proposed using metric learning to regularize the latent representations of different classes. Specifically, a triplet loss is utilized, in which latent representations of the clean image, its corresponding AE and an image from another class are considered as the {\em positive}, {\em anchor}, and {\em negative} example, respectively. However, this strategy only considers the {\em one-directional} adversarial attack, \ie, from the clean image to its adversarial example, making it less efficient. 

To address the above issue, we propose a \textit{Bi-directional attack Metric Learning} (BML) to provide a more efficient and strong constraint. Specifically, the original clean image ({\em positive}) is first attacked to its {\em most confusing class}, which is the class that has the smallest loss other than the correct label, to get the {\em forward} adversarial example ({\em anchor}). Then, a clean image is randomly picked from the most confusing class and is attacked to the original image to get the {\em backward} adversarial example as the {\em negative}. 

By integrating AGKD and BML, our \textbf{AGKD-BML} model outperforms the state-of-the-art models on two widely used datasets, CIFAR-10 and SVHN, under different attacks. In summary, our contribution is three-fold:
\begin{itemize}
\vspace{-1.61mm}\item An attention guided knowledge distillation module is proposed to transfer attention information of clean image to the student model, such that the intermediate features corrupted by adversarial examples can be corrected.

\vspace{-1.61mm}\item A bidirectional metric learning is proposed to efficiently constrain the representations of the different classes in feature space, by explicitly shortening the distance between original image and its forward adversarial example, while enlarging the distance between the forward adversarial example and the backward adversarial example from another class.

\vspace{-1.61mm}\item We conduct extensive adversarial robustness experiments on the widely used datasets under different attacks, the proposed AGKD-BML model outperforms the state-of-the-art approaches with both the qualitative (visualization) and quantitative evidence.
\end{itemize}

\section{Related Works}

\noindent \textbf{Adversarial Attacks.} Generally, there are two types of adversarial attacks: {\em white-box} attack where the adversary has full access to the target model, including the model parameters, and the {\em black-box} attack, where the adversary has almost no knowledge of the target model. For white-box attack, Szegedy \etal~\cite{szegedy2013intriguing} discovered the vulnerability of deep networks against adversarial attacks. They used a box-constrained L-BFGS method to generate effective adversarial attacks. After that, several algorithms were developed to generate adversarial examples. As a one-step attack, the fast gradient sign method (FGSM) proposed in~\cite{goodfellow2014explaining} uses the sign of the gradient to generate attacks, with $\ell_{\infty}$-norm bound. In~\cite{kurakin2016adversarial}, Kurakin~\etal extended FGSM by applying it iteratively and designed basic iterative method (BIM). A variant of BIM was proposed in~\cite{dong2018boosting} by integrating momentum into it. DeepFool~\cite{moosavi2016deepfool} tried to find the minimal perturbations based on the distance to a hyperplane and quantify the robustness of classifiers. In~\cite{papernot2016limitations}, the authors introduced a Jacobian-based Saliency Map Attack. The projected gradient descent (PGD) was proposed in~\cite{madry2017towards} as a multi-step attack method. The CW attack, a margin-based attack, was proposed in~\cite{carlini2017towards}. Recently, Croce~\etal introduced a parameter-free attack named AutoAttack~\cite{AA_atack_2020}, which is an ensemble of two proposed parameter-free versions of PGD attacks and the other two complementary attacks, \ie, FAB~\cite{croce2020minimally} and Square Attack~\cite{andriushchenko2020square}. It evaluates each sample based on its worst case over these four diverse attacks which includes both white-box and black-box ones.
Besides the additive attacks,~\cite{fawzi2015manitest, goodfellow2009measuring, kanbak2018geometric} show that even small geometric transformations, such as affine or projective transformation can fool a classifier. In addition to those attacks on the image input to the model, attempts are made to design adversarial patches that can fool the model in the physical world~\cite{eykholt2018robust, huang2020universal, kurakin2016adversarial}. On the other side of the coin, adversarial attacks may also be used to improve the model performance~\cite{xie2020adversarial, PASM21, pan2020adversarial}.

\noindent \textbf{Adversarial defense.} Adversarial training-based models, which aim to minimize the classification loss to the strongest adversarial examples (maximal loss within a $\epsilon-$ball), are believed as one of the most effective and widely used defense methods. In practice, they iteratively generate adversarial examples for training. In~\cite{goodfellow2014explaining}, Goodfellow \etal generated the adversarial examples by FGSM, while Madry \etal~\cite{madry2017towards} used the Projected Gradient Descent (PGD) attacks during adversarial training. Many variants based on adversarial training were proposed in recent years. For example, \cite{shafahi2019adversarial} computed the gradient for attacks and the gradient of model parameters at the same time, and significantly reduced the computation time. Adversarial logit paring~\cite{kannan2018adversarial} constraints distance between the logits from a clean image and its adversarial example, while~\cite{Mao2019} and~\cite{Zhong2019} built a triplet loss between a clean image, its corresponding adversarial example and a negative sample. 
TRADES~\cite{zhang2019theoretically} optimized the trade-off between robustness and accuracy. In~\cite{Wang2019}, the authors designed an adversarial training strategy with both adversarial images and adversarial labels. In~\cite{zhang2019defense}, feature scattering is used in the latent space to generate adversarial examples and further improved the model’s accuracy under different attacks. Xie \etal~\cite{xie2019feature} proposed feature denoising models by adding denoise blocks into the architecture to defend the attack.

Most of the existing adversarial training-based models focus on the on-training model that utilizes adversarial examples, which may be corrupted, but have not explored the information from the model trained on clean images.

\textbf{Other adversarial defense models.} 
In~\cite{lu2017safetynet, xu2017feature}, the authors proposed to firstly detect and reject adversarial examples. Several methods proposed to estimate the clean image by using a generative model~\cite{song2018pixeldefend, sun2019adversarial, yuan2020ensemble}. Cohen \etal~\cite{cohen2019certified} proposed to use randomized smoothing to improve adversarial robustness. 
There are also several works utilized large scale external unlabeled data to improve the adversarial robustness, \eg,~\cite{carmon2019unlabeled} and~\cite{uesato2019labels}.

In this paper, we focus on improving the adversarial robustness of the model itself without using external data or pre-processing the testing data.

\section{Proposed Method}

\begin{figure*}[htbp]
  \centering
  \includegraphics[width=.9\linewidth]{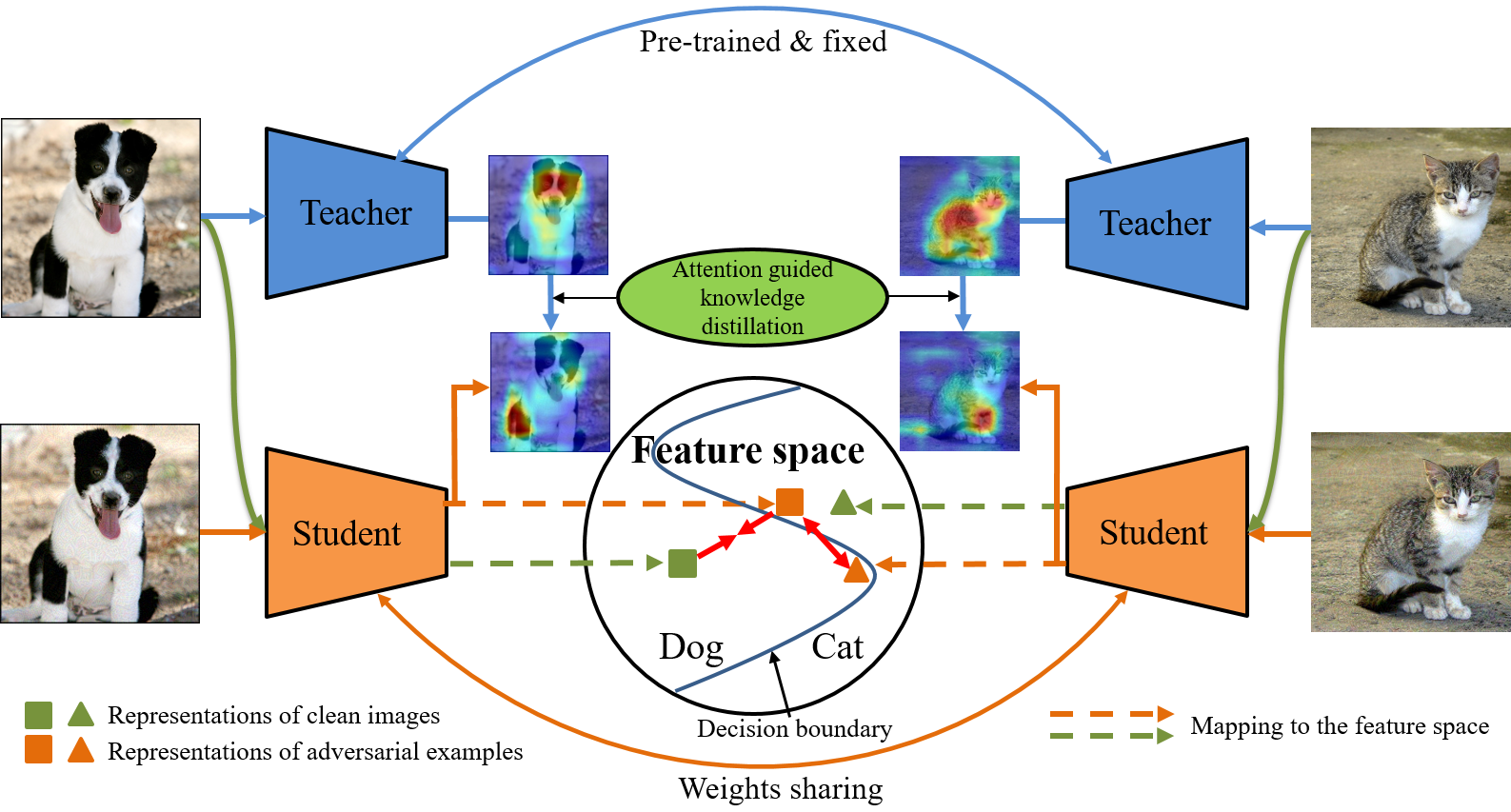}
  \caption{The framework of the proposed AGKD-BML model. Top-left is a clean image that belongs to ``Dog", and bottom-left is its adversarial example (AE) targeted attack to its most confusing class ``Cat". Similarly, top-right and bottom-right are a clean image of ``Cat" and its AE targeted ``Dog", respectively. ``Teacher" is the model pre-trained on clean images and ``Student" is the on-training model. The AE will fool the model by 1) focusing on incorrect regions, and 2) crossing the decision boundary in feature space. Attention guided knowledge distillation, illustrated as a green ellipse, is used for correcting the focus region. Bidirectional metric learning, illustrated as red arrows in the ``Feature space", is used to pull the AEs back to their original classes. Better viewed in color.}
  \label{fig:Framework}
\end{figure*}

In this section, we present the framework of our proposed AGKD-BML model in detail. As illustrated in Figure~\ref{fig:Framework}, AGKD-BML framework consists of two modules, \ie, the attention guided knowledge distillation (AGKD) module and the bidirectional attack metric learning (BML) module. The AGKD module is used for distilling attention knowledge of the clean image to the student model, to obtain a better attention map for adversarial example, as well as correcting the corrupted intermediate features. The BML module efficiently regularizes the representation in feature space by using bidirectional metric learning. In the rest of this section, we first briefly introduce the standard adversarial training (AT) and (non-)targeted adversarial attack, and then describe the two modules of our proposed model and the integration of them.

\subsection{Preliminaries}
We first briefly describe the standard adversarial training (AT)~\cite{madry2017towards}. Suppose we have a labeled $C$-class classification dataset $\mathcal{D}=\left\{(x, y)\right\}_{i=1}^N$ of $N$ samples, where the label $y\in\left\{1,2,\dots,C\right\}$. There are two types of adversarial attacks, \ie, the non-targeted attacks and the targeted attacks, which can be formulated as eq.~(\ref{eq:non-targeted}) and~(\ref{eq:targeted}), respectively:
\begin{align}
    \max \limits_{\delta \in\Delta} \mathcal{L}(f_{\theta}(x+\delta), y) \label{eq:non-targeted} \\
    \min \limits_{\delta \in\Delta} \mathcal{L}(f_{\theta}(x+\delta), y_t) \label{eq:targeted}
\end{align}
where $\delta$ is the perturbation added to the image $x$, $\Delta$ provides an $\ell_{\infty}$-norm bound of the perturbation, and $f_{\theta}(\cdot)$ and $\mathcal{L}(\cdot)$ to denote the network with model parameters $\theta$ and the loss function, respectively. Non-targeted attacks maximize the loss function given the correct label $y$, while targeted attacks minimize the loss function given the target label $y_t$.

Standard AT uses non-targeted PGD (Projected Gradient Descent) attack~\cite{madry2017towards} during training, which can be formulated as a min-max optimization problem:
\begin{equation}\label{eq:mimmax} 
\min \limits_{\theta} \mathop{\mathbb{E}} \limits_{x\in\mathcal{D}} \left[\max \limits_{\delta \in\Delta} \mathcal{L}(f_{\theta}(x+\delta), y)\right]
\end{equation}
In the objective function, the outer minimization is the update of the model parameters while the inner maximization is for generating adversarial attacks. Specifically, PGD is used to generate attacks, which is an iterative non-targeted attack with random start at the beginning. In this paper, following~\cite{Wang2019}, we use targeted attacks during training with the most confusing class as target class.

\subsection{Attention guided knowledge distillation}

To distill attention information of clean images to the student model, we propose an attention guided knowledge distillation module. Figure~\ref{fig:attention} shows attention maps of a clean image (``German shepherd”) and its adversarial example (``Planetarium”). As a class relevant attention map, the Grad-CAM~\cite{selvaraju2017grad} shows the focusing region related to a specific class. 
From the figure we can see that although the adversarial example degrades the attention map of original class, it hurts the attention map of target (incorrect) class much more largely, and makes the features of incorrect class overwhelm the correct class and dominate overall (as such makes the model mis-classified). We argue that only distilling the class relevant attention information has limited effects on correcting the features of the targeted class. Therefore, we propose to distill class irrelevant attention information (see section~\ref{sec:attention}) of clean images. We provide more explanations and discussions to justify our choice in \textbf{\em supplementary materials.}

\subsubsection{Class irrelevant attention map}\label{sec:attention}

We generate the class irrelevant attention map at the last convolutional layer. Specifically, we treat the backbone neural network until the last convolutional layer as a feature extractor, denoted by $\mathcal{F}(x)$ for a given image $x$, where $\mathcal{F}(x)\in\mathbb{R}^{C\times H \times W}$. We then produce an operator, denoted by $\mathcal{A}(\cdot)$, to map the feature map to the two-dimensional attention map, $\mathcal{A}(\mathcal{F}(x))\in\mathbb{R}^{1\times H \times W}$. In this paper, we simply pick the average pooling through the channel dimension (or identical weights 1$\times$1 convolution) as $\mathcal{A}$.

\subsubsection{Knowledge distillation}

The knowledge distillation (KD)~\cite{hinton2015distilling} utilizes a {\em student-teacher (S-T) learning framework} to transfer information learned from the teacher model to the student model. In this paper, we treat the model trained on the natural clean images by standard training as the teacher model and the one under adversarial training as the student model. The attention information is what we expect to transfer from the teacher model to the student model. As the teacher model is trained on the clean images with high testing accuracy, it is able to provide correct regions that model should focus on. Therefore, the attention map of the clean image extracted by the teacher model will transfer to the student model. The loss function of this attention guided knowledge distillation is written as:
\begin{equation}
    \mathcal{L}_{KD}(x_t, x_s)=D(\mathcal{A}(\mathcal{F}_t(x_t)), \mathcal{A}(\mathcal{F}_s(x_s)))  \label{eq:KD}
\end{equation}
where $x_t$ and $x_s$ are input images of the teacher and student model, respectively, and $\mathcal{F}_t$ and $\mathcal{F}_s$ are feature extractors of the teacher and student models, respectively. $D(\cdot)$ is the distance function (\eg, $\ell_1$) to measure the similarity between these two attention maps. Given an adversarial example, the AGKD guides the student model to focus on the same regions as its clean image.

\subsection{Bidirectional attack metric learning}

In our work, we use the targeted attack to obtain the adversarial examples. Let $x^{s}$ refers to a samples with the label $y=s$, and $x^{s}_{t}$ refers to an adversarial example of $x^{s}$ with the target label $y_t=t$. In this paper, the forward adversarial example is targeted towards the most confusing class, which is defined as follow: 
\begin{equation}
    y_{mc} = \argmin_{y\neq{s}}\mathcal{L}(f(x^{s}), y).  
\label{eq:MC}
\end{equation}

Given an original clean image $x^s$, we first generate the targeted adversarial example $x^{s}_{y_{mc}}$ towards its most confusing class. Then, we randomly select a sample $x^{y_{mc}}$ from the most confusing class, and generate its adversarial example $x^{y_{mc}}_{s}$ that targeted back to the original label $s$. We utilize $x^{s}$, $x^{s}_{y_{mc}}$, $x^{y_{mc}}_{s}$ as positive, anchor and negative sample, respectively. The triplet loss is defined as:
\begin{equation} \label{eq:triplet}
    \begin{split}
    & \mathcal{L}_{tr}(x_a, x_p, x_n) \\
    & = [d(E(x_a),E(x_p))-d(E(x_a),E(x_n))+m]_+,
    \end{split}
\end{equation}
where $x_a, x_p, x_n$ denote to positive, anchor and negative samples, respectively. $E(\cdot)$ is the representation from the penultimate layer of the model. $d(a, b)$ denotes the distance between two embeddings $a$ and $b$, which is defined as the angular distance $d(a, b) = 1-\frac{\left|a \cdot b \right|}{\|a\|_2 \|b\|_2}$, following~\cite{Mao2019}. $m$ is the margin. Comparing to the previous metric learning based adversarial training, \eg,~\cite{Mao2019} and~\cite{Zhong2019} which only consider forward adversarial example, we consider both the forward and backward adversarial examples. Therefore, we name it the bidirectional metric learning.

By adding a $\ell_2$-norm regularization on the embedding, the final BML loss function is written as:
\begin{equation}\label{eq:metric} 
\mathcal{L}_{BML}=\lambda_1\mathcal{L}_{tr}(x^{s}_{y_{mc}}, x^{s}, x^{y_{mc}}_{s}) + \lambda_2\mathcal{L}_{norm},
\end{equation}
where $\mathcal{L}_{norm}= \|E(x^{s}_{y_{mc}})\|_2 + \|E(x^{s})\|_2 +\|E(x^{y_{mc}}_{s})\|_2$ is the normalization term, and $\lambda_1$ and $\lambda_2$ are the trade-off weights for the two losses.

\subsection{Integration of two modules}
We integrate the attention guided knowledge distillation and bidirectional metric learning together to take the benefits from both modules. As we consider bidirectional adversarial attack, we have two clean/adversarial image pairs, $x^{s}$/$x^{s}_{y_{mc}}$ and $x^{y_{mc}}$/$x^{y_{mc}}_{s}$. For both pairs, we apply the AGKD from the attention map of the clean image obtained by teacher model to the student model, which can be formulated as:
\begin{equation}
\label{eq:AGKD}
\begin{split}
    \mathcal{L}_{AGKD} =& \mathcal{L}_{KD}(x^s, x^s_{y_{mc}}) + \mathcal{L}_{KD}(x^{y_{mc}}, x^{y_{mc}}_s)\\
\end{split}
\end{equation}
where the first term denotes the AGKD loss for the forward attack pair, \ie, $x^{s}$ and $x^{s}_{y_{mc}}$, while the second term denotes the backward attack pair, \ie, $x^{y_{mc}}$ and $x^{y_{mc}}_{s}$, 

By combining the standard cross entropy loss used in the traditional adversarial training, the BML loss, and the AGKD loss, the final total loss is:
\begin{equation}
    \mathcal{L}_{total}=\mathcal{L}_{ce} +\mathcal{L}_{AGKD} + \mathcal{L}_{BML}  \label{eq:total}
\end{equation}
The overall procedure of AGKD-BML model is shown in Algorithm.~\ref{alg:alg}.

\begin{algorithm}
  \caption{AGKD-BML model}
  \label{alg:alg}
  \KwIn{Clean image set $\mathcal{D}$, epoch number $N$, batch size $b$, learning rate $\gamma$}

  \KwOut{Network parameter $\theta$}
  \For{{\em epoch} $= 1, ..., N$}
  {
    \For{{\em minibatch} $\{x_i, y_i\}^b_{i=1}$ } 
    {
      initialize $\mathcal{L}_{batch}=0$ \;
      \For{{\em sample one} $x^s$ {\em belongs to class} $s$ } 
      {
        a. find its MC class $y_{mc}$ by Eq.~\ref{eq:MC}, sample one data $x^{y_{mc}}$ from class $y_{mc}$ \;
        b1. obtain $x^s_{y_{mc}}$ by attacking $x^s$ to $y_{mc}$ \;  
        b2. obtain $x_s^{y_{mc}}$ by attacking $x^{y_{mc}}$ to $s$ \; 
        c. calculate $\mathcal{L}_{BML}$ by Eq.~\ref{eq:metric} \;
        d. calculate $\mathcal{L}_{AGKD}$ by Eq.~\ref{eq:AGKD} \;
        e. calculate $\mathcal{L}_{total}$  by Eq.~\ref{eq:total} \;
        f. update $\mathcal{L}_{batch}=\mathcal{L}_{batch} + \frac{1}{b}\mathcal{L}_{total}$~.
      }
      update $\theta= \theta - \gamma\cdot  \nabla_{\theta}\mathcal{L}_{batch}$~.
    }
  }
   return  $\theta$\;
\end{algorithm}

\section{Experiments}

\subsection{Experimental settings}

\textbf{Dataset} We evaluate our method on two popular datasets: CIFAR-10 and SVHN. \textit{CIFAR-10} consists of 60k 3-channel color images with size of $32\times32$ in 10 classes, in which 50k images for training and 10k images for testing. \textit{SVHN} is the street view house number dataset, which has 73257 images for training and 26032 images for testing. We evaluate model on a larger datasets: Tiny ImageNet, and the results are shown in \textbf{\em supplementary materials.}


\textbf{Comparison methods} We use comparison methods include: (1) {\em undefended model (UM)}, where the model is trained by standard training; (2) {\em adversarial training (AT)}~\cite{madry2017towards}, which uses non-targeted PGD adversarial examples (AEs) for training; 
(3) {\em single-directional metric learning (SML)}~\cite{Mao2019}; (4) {\em Bilateral}~\cite{Wang2019}, which generates AEs on both images and labels; (5) {\em feature scattering (FS)}~\cite{zhang2019defense}, where adversarial attacks for training are generated with feature scattering in the latent space; (6) and (7) utilize the {\em channel-wise activation suppressing (CAS)}~\cite{CAS_2021} on TRADES~\cite{TRADES_2019} and MART~\cite{MART_2020}, respectively, which showed the superior compared to the original version. Note that Bilateral, FS generate AEs by using single-step attacks {\em in training}, while AGKD-BML uses 2-steps attacks, ``AT" and ``SML" use 7-step attacks, and ``TRADES+CAS" and ``MART+CAS" use 10-step attacks. To fairly compare to these multi-step attack models, we also train a 7-step attack variant of AGKD-BML, referred as to ``AGKD-BML-7". We test the models with various attacks including FGSM~\cite{goodfellow2014explaining}, BIM~\cite{kurakin2016adversarial}, PGD~\cite{madry2017towards}, CW~\cite{carlini2017towards}, MIM~\cite{dong2018boosting} with different attack iterations. We also evaluate the models in a per-sample manner using AutoAttack (AA)~\cite{AA_atack_2020}, which is an ensemble of four diverse attacks. Finally, we also test the black-box adversarial robustness of the model.

\textbf{Implementation details} Following~\cite{madry2017towards} and~\cite{Mao2019}, we use Wide-ResNet (WRN-28-10)~\cite{ZagoruykoK16}, 
and set the initial learning rate $\gamma$ as 0.1 for CIFAR-10 and 0.01 for SVHN. We use the same learning rate decay points as~\cite{Wang2019} and~\cite{zhang2019defense}, where decay schedule $\left[100, 150\right]$ for CIFAR-10 and $\left[60, 90\right]$ for SVHN, with 200 epochs in total. ``AGKD-BML-7" has the learning rate that decays at 150 epochs and the training stops at 155 epochs, following the suggestions in~\cite{MART_2020, rice2020overfitting}. 
In training phase, the perturbation budget $\epsilon=8$ and label smoothing equals to 0.5 following~\cite{zhang2019defense}. In the AGKD module, we adopt $\ell_1$ norm to measure the similarity between attention maps. For the BML module, parameters are the same as~\cite{Mao2019}, \ie, margin $m=0.03$, $\lambda_1=2$ and $\lambda_2=0.001$. 

\subsection{Evaluation of adversarial robustness}

\begin{table*}[htbp]\small
\centering
\caption{\label{tab:full_result} Evaluation results on CIFAR-10 and SVHN, under different widely used attacks. For CIFAR-10 dataset, we grouped the models by small-number or larger-number steps attack {\em in training}. ``Bilateral" and ``FS" use one-step attack, AGKD-BML uses two-step attack, and others use large-number steps with the step numbers show followed by the model names. ``AGKD-BML-7" is a veriant of AGKD-BML that uses 7-step attack for training. The best accuracy for each group is illustrated as bold, and the overall best accuracy is highlighted.
}
\begin{tabular}{c|cccccccc}
\hline \hline
\multicolumn{9}{c}{CIFAR-10} \\
\hline
Attacks(steps) & clean & FGSM & BIM(7)  & PGD (20) & PGD (100) & CW (20) & CW (100) & AA~\cite{AA_atack_2020} \\
\hline 
UM & \colorbox{gray}{\textbf{95.99$\%$}}  & 31.39$\%$  & 0.38$\%$  & 0$\%$  & 0$\%$  & 0$\%$  & 0$\%$  & 0$\%$ \\
\hline
Bilateral~\cite{Wang2019} & 91.2$\%$  & 70.7$\%$  & -  & 57.5$\%$  & 55.2$\%$  & 56.2$\%$  & 53.8$\%$  & 29.35$\%$  \\
FS~\cite{zhang2019defense} & 90.0$\%$  & \colorbox{gray}{\textbf{78.4$\%$}}  & - & 70.5$\%$  & 68.6$\%$  & 62.4$\%$  & 60.6$\%$  & 36.64$\%$  \\
AGKD-BML  & \textbf{91.99$\%$}  & 76.69$\%$  & \colorbox{gray}{\textbf{73.81$\%$}}  & \colorbox{gray}{\textbf{71.02$\%$}}  & \colorbox{gray}{\textbf{70.72$\%$}}  & \textbf{63.67$\%$}  & \colorbox{gray}{\textbf{62.55$\%$}}  & \textbf{37.07$\%$} \\
\hline
AT-7~\cite{madry2017towards} & 86.19$\%$  & 62.42$\%$  & 54.99$\%$  & 45.57$\%$  & 45.22$\%$  & 46.26$\%$  & 46.05$\%$  & 44.04$\%$ \\
SML-7~\cite{Mao2019} & 86.21$\%$  & 58.88$\%$  & 52.60$\%$  & 51.59$\%$  & 46.62$\%$  & 48.05$\%$  & 47.39$\%$  & 47.41$\%$ \\
TRADES+CAS-10~\cite{CAS_2021} & 85.83$\%$  & 65.21$\%$  & - & 55.99$\%$  & -  & \colorbox{gray}{\textbf{67.17$\%$}}  & -  & 48.40$\%$  \\
MART+CAS-10~\cite{CAS_2021} & \textbf{86.95$\%$}  & 63.64$\%$  & - & 54.37$\%$  & -  & 63.16$\%$ & -  & 48.45$\%$  \\
AGKD-BML-7 & 86.25$\%$  & \textbf{70.06$\%$}  & \textbf{64.97$\%$}  & \textbf{57.30$\%$}  & \textbf{56.88$\%$} & 53.36$\%$ & \textbf{52.95$\%$} & \colorbox{gray}{\textbf{50.59$\%$}} \\
\hline \hline
\multicolumn{9}{c}{SVHN} \\
\hline
 Attacks(steps) & clean & FGSM & BIM(10)  & PGD (20) & PGD (100) & CW (20) & CW (100) & MIM (40) \\
\hline 
UM & \textbf{96.36$\%$}  & 46.33$\%$  & 1.54 $\%$  & 0.33$\%$  & 0.22$\%$  & 0.37$\%$  & 0.24$\%$ & 5.39$\%$ \\
Bilateral~\cite{Wang2019} & 94.1$\%$  & 69.8$\%$  & -  & 53.9$\%$  & 50.3$\%$  & -  & 48.9$\%$ & - \\
FS~\cite{zhang2019defense} & 96.2$\%$  & 83.5$\%$  & -  & 62.9$\%$  & 52.0$\%$  & 61.3$\%$  & 50.8$\%$  & - \\
AT-7~\cite{madry2017towards} & 91.55$\%$  & 67.13$\%$  & 54.03$\%$  & 45.64$\%$  & 44.02$\%$  & 47.14$\%$  & 45.66$\%$ & 52.13$\%$ \\
SML-7~\cite{Mao2019} & 83.95$\%$  & 70.28$\%$  & 57.58$\%$  & 51.91$\%$  & 49.81$\%$  & 51.25$\%$  & 49.31$\%$ & 43.80$\%$ \\
TRADES+CAS-10~\cite{CAS_2021} & 91.69$\%$  & 70.79$\%$  & - & 55.26$\%$  & -  & 60.10$\%$  & -  & -  \\
MART+CAS-10~\cite{CAS_2021} & 93.05$\%$  & 70.30$\%$  & - & 51.57$\%$  & -  & 53.38$\%$  & -  & -  \\
\hline
AGKD-BML & 95.04$\%$  & \textbf{89.32$\%$}  & \textbf{75.06$\%$}  & \textbf{74.94$\%$}  & \textbf{69.23$\%$}  & \textbf{69.85$\%$}  & \textbf{62.22$\%$} & \textbf{76.86$\%$} \\
\hline \hline

\end{tabular}
\end{table*}

We evaluate our model's adversarial robustness and report the comparisons in Table~\ref{tab:full_result}. 
The results on ``clean" images are used as a baseline for evaluating how much the accuracy of the defenders will drop as increasing the adversarial robustness. It is shown in Table~\ref{tab:full_result}, AGKD-BML overall outperforms the comparison methods on CIFAR-10. AGKD-BML also shows better adversarial robustness on SVHN dataset with a large margin.

Interestingly, in Table~\ref{tab:full_result}, we observed that AGKD-BML showed different superiors to different attacks, \ie, AGKD-BML trained on 7-step attack has higher performance than that trained on 2-step attack against AA, but much lower performance against the regular attacks, \eg PGD and CW. The reason for this phenomenon is, in our opinion, that compared to the regular attacks, AA is an ensemble of four different types of attack, including white-box and black-box ones, which requires the generalization capability of defense against different types of attack. The generation of the 7-step attack significantly increases the diversity of AEs used for training and thus, it improves the robustness against AA with some sacrifice on accuracy against regular attacks. On the other hand, the generation of the 2-step attack focuses more on the regular attacks but less diverse, which makes it has lower performance against AA. As an empirical defense method, we argue the model trained by small-number-step attack is still useful in some scenarios that the adversarial attacks are known. We provide more results of AGKD-BML model trained on large-number-step attacks against AA in \textbf{\em supplementary materials.}

\subsection{Ablation study}

\begin{table}[htbp]
\small
\caption{\label{tab:ab} Ablation study on CIFAR-10 dataset.}
\centering
\begin{tabular}{c|ccc}
\hline \hline
& FGSM  & PGD (20)  & CW (20) \\
\hline 
UM & 31.39$\%$  & 0$\%$  & 0$\%$  \\
AT~\cite{madry2017towards} & 62.42$\%$  & 45.57$\%$  & 46.26$\%$  \\
SML~\cite{Mao2019} & 58.88$\%$  & 51.59$\%$  & 48.05$\%$  \\
BML & 71.08$\%$  & 60.51$\%$  & 56.53$\%$ \\
AGKD & 75.57$\%$  & 65.93$\%$  & 60.71$\%$  \\
\hline
AGKD-BML & 76.69$\%$  & 71.02$\%$  & 63.67$\%$ \\
\hline \hline
\end{tabular}
\end{table}

\begin{figure*}[htbp]
  \centering
  \includegraphics[width=.995\linewidth]{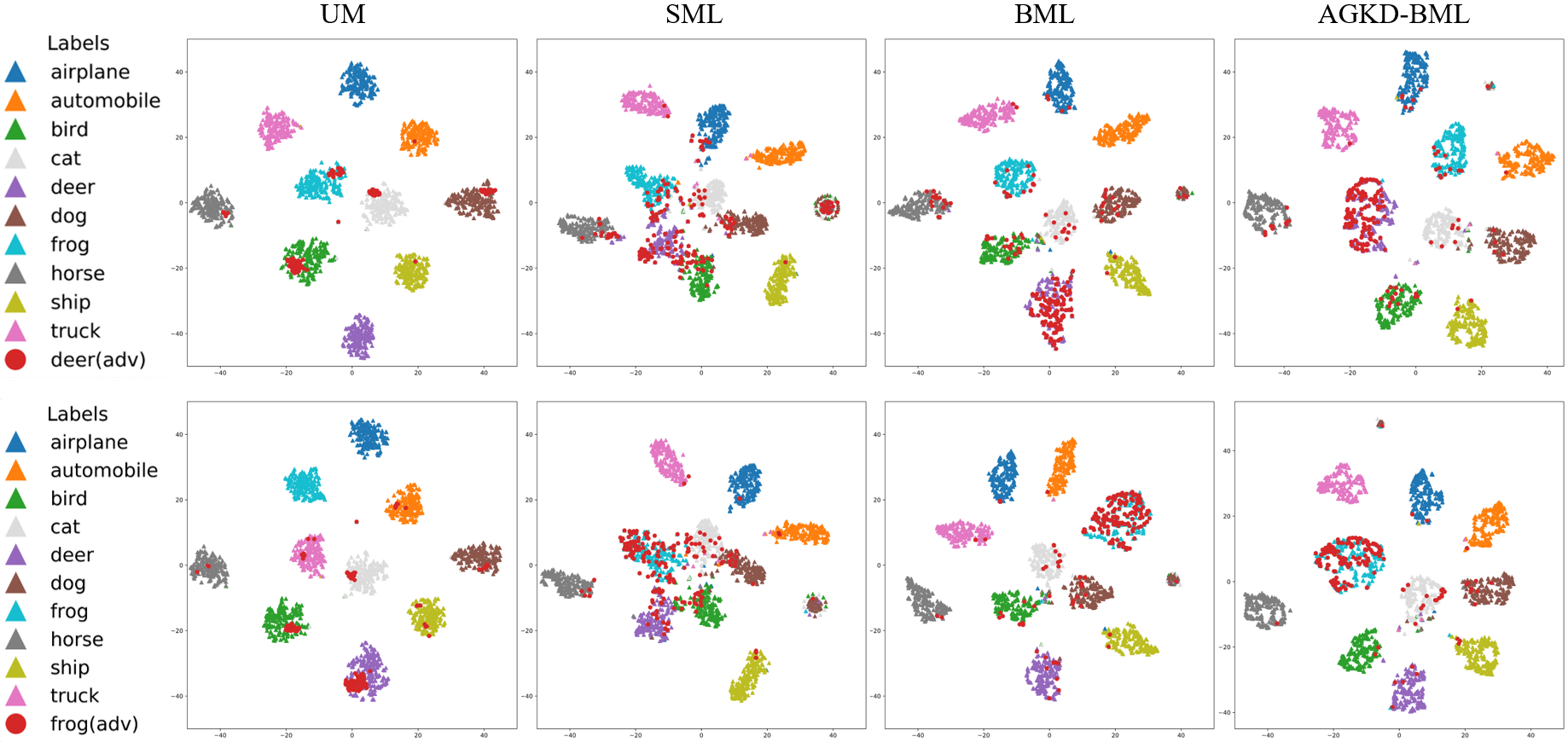}
  \caption{t-SNE plots for illustrating the sample representations in feature space. The triangle points with different colors represent the clean images in different classes, and the red circle points are the adversarial examples under PGD-20 attack. Best viewed in color.}
  \label{fig:tsne}
\end{figure*}

We analyze the ablation effect of each component of AGDK-BML on CIFAR-10 dataset. The quantitative and qualitative results are shown in Table~\ref{tab:ab} and Figure~\ref{fig:tsne}, respectively. ``UM", ``AT" and ``SML" are the same models described above. ``BML" denotes the {\em bidirectional} metric learning without using any knowledge distillation. ``AGKD" denotes the model applied attention map guided knowledge distillation without any metric learning. 
In Figure~\ref{fig:tsne}, we provide the t-SNE plots to show the sample representations in feature space. The triangle points with different colors represent the clean images in different classes, while the red circle points are the AEs under PGD-20 attack. We show AEs from two classes (\ie, deer and frog). 

``UM" shows how the adversarial attacks behave if a model dose not have any defense. A simple one step attack FGSM drops UM's accuracy to $\sim$30$\%$, while the multi-step attacks, \eg, PGD-20 and CW-20, drop its accuracy to 0$\%$. It is also visualized in the first column of Figure~\ref{fig:tsne}, where all the AEs locate far from their original class, and fit into the distributions of other classes. As a standard benchmark defense model, ``AT" provides a baseline for improvements on both single-step and multi-step attacks. 

\textbf{The Effect of Bidirectional Metric Learning} ``SML" and ``BML" both apply metric learning to constrain the clean image and its AE to keep a short distance, while push away the images in different classes. The difference between them is that SML only considers forward attacks and BML considers both forward and backward attacks. In the second and third columns of Figure~\ref{fig:tsne}, we can see that the SML does pull many of the AEs back to their original class, i.e., purple in the first row and cyan in the second row. However, one of the side effects of SML is that it makes classes confusing for clean images and thus may make a significant accuracy drop on clean images. In contrast, BML keeps better separations between different classes, and has much less amount of AEs located far away compared to SML. It demonstrates the benefit of the bidirectional strategy.

\begin{figure}[htbp]
  \centering
  \includegraphics[width=.995\linewidth]{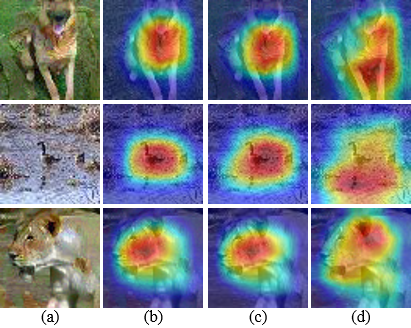}
  \caption{(a) Sample adversarial examples. (b) Attention maps obtained from clean images, which are treated as ground truth. (c) and (d) are the attention maps obtained by AGDK-BML and AT.}
  \label{fig:heatmap}
\end{figure}

\textbf{The Effect of Attention Guided Knowledge Distillation} Utilizing ``AGKD" alone is able to obtain a good accuracy, which is better than BML. By integrating both AGKD and BML, the proposed AGKD-BML obtain the best performance in terms of both quantitative and qualitative results. In the fourth column of Figure~\ref{fig:tsne}, AGKD-BML pulls most of the AEs back to their original class, while keeps better separation between classes than BML does. We also provide the attention maps of the AEs obtained by the trained models in Figure~\ref{fig:heatmap}. Compared to AT, AGDK-BML obtains better attention maps which are more identical to the ones of clean images. This suggests that the AGKD does help on correcting the representation of AEs in feature space.

\begin{figure}[htbp]
  \centering
  \includegraphics[width=.998\linewidth,height=.418\linewidth]{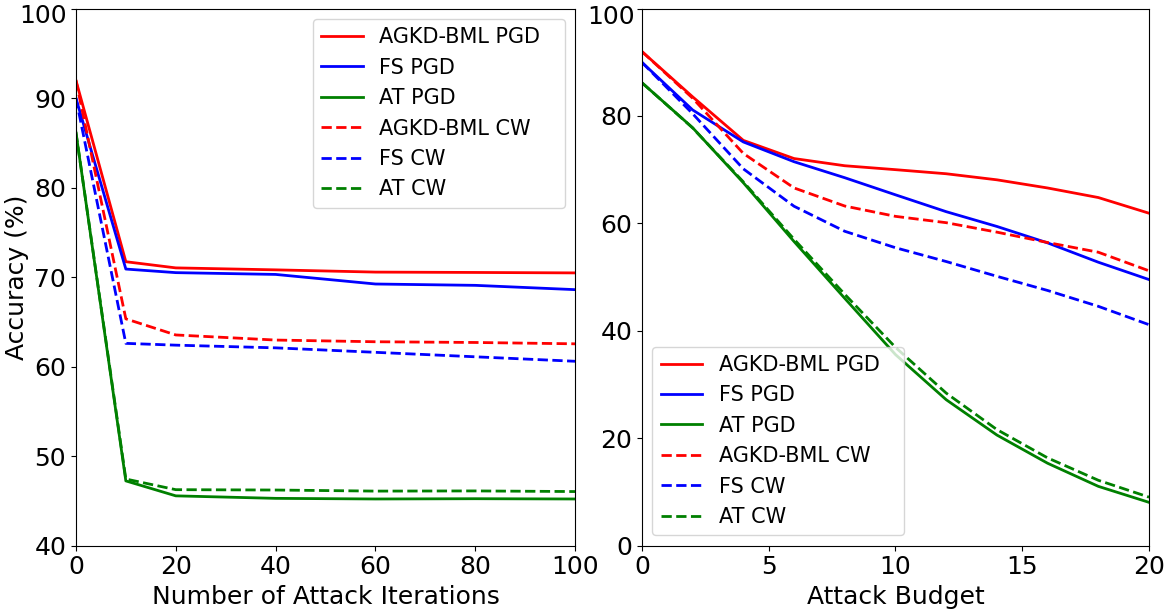}
  \caption{The accuracy under different attack iterations and perturbation budgets ($\epsilon$).}
  \label{fig:budget}
\end{figure}

\textbf{Different attack iterations and budgets} We evaluate model robustness under different PGD attack iterations, and different attack budgets ($\epsilon$) with a fixed attack iteration of 20. It is shown in Figure~\ref{fig:budget} that AGKD-BML consistently outperforms two comparison methods, \ie, feature scatter (FS)~\cite{zhang2019defense} and standard AT, on all numbers of attack iterations up to 100 and all attack budgets up to $\epsilon=20$. Moreover, AGKD-BML also shows more robust to large attack budgets as the accuracy drops are significantly less than the other two comparison methods.

\subsection{Black box adversarial robustness}
To evaluate the black-box adversarial robustness, i.e., the adversary has no knowledge about the model,  we generate an AE for each clean image in CIFAR-10 testing set by using natural models under PGD-20 attack with $\epsilon=8$. Then the AGKD-BML model, as well as the comparison models, are tested on the generated adversarial example data. As demonstrated in Table~\ref{tab:blackbox}, AGKD-BML model achieves the best accuracy among the models suggesting that AGKD-BML is robust to the black-box attacks as well.

\begin{table}[!t]
\small
\centering
\caption{\label{tab:blackbox} Black-box adversarial robustness.}
\begin{tabular}
{@{\hspace{1.5mm}}c@{\hspace{1.5mm}} @{\hspace{1.5mm}}c@{\hspace{1.5mm}} @{\hspace{1.5mm}}c@{\hspace{1.5mm}} @{\hspace{1.5mm}}c@{\hspace{1.5mm}} @{\hspace{1.5mm}}c@{\hspace{1.5mm}} }
\hline \hline
 AT~\cite{madry2017towards}  &  FS~\cite{zhang2019defense} & Bilateral~\cite{Wang2019} & SML~\cite{Mao2019} & AGKD-BML\\
\hline 
 85.4$\%$ & 88.9$\%$ & 89.9$\%$ & 86.4$\%$ & \textbf{90.75$\%$}  \\
\hline \hline
\end{tabular}
\end{table}


\subsection{Discussion}

Based on the analysis in~\cite{athalye2018obfuscated}, we claim that the robustness of our model is not from gradient obfuscation for the following reasons: 1) In table~\ref{tab:full_result}, iterative attacks are stronger than one-step attack (FGSM). 2) Figure~\ref{fig:budget} shows that the accuracy monotonically declines under attacks with more steps or increasing perturbation budgets. 3) Table~\ref{tab:blackbox} shows that black-box attacks have a lower success rate (higher accuracy) than white-box attacks. 4) We evaluate our model against a gradient-free attack~\cite{uesato2018adversarial} and the accuracy is 88.67$\%$, which is higher than gradient-based attacks (71.02$\%$ for PGD20).


\section{Conclusion}
We proposed a novel adversarial training based model, named as \textit{AGKD-BML}, that integrates two modules, \ie, the attention guided knowledge distillation module and the bi-directional metric learning module. The first module transfers attention knowledge of the clean image from a teacher model to a student model, so as to guide student model for obtaining better attention map, as well as correcting the intermediate features corrupted by adversarial examples. The second module efficiently regularizes the representation in the feature space, by shortening the representation distance between original image and its forward adversarial example, while enlarging the distance between the forward and backward adversarial examples. Extensive adversarial robustness experiments on two popular datasets with various attacks show that our proposed AGKD-BML model consistently outperforms the state-of-the-art approaches.

\vspace{.06in}
\noindent\textbf{Acknowledgement.} This work is supported by the U.S. Department of Energy, Office of Science, High Energy Physics under Award Number DE-SC-0012704 and the Brookhaven National Laboratory LDRD \#19-014, and in part by National Science Foundation Award IIS-2006665.


\newpage
\appendix
\begin{center} \noindent\textbf{\LARGE Appendix} \end{center} 

\section{Evaluation on Tiny ImageNet}
We evaluate our method on a larger dataset, \ie, \textit{Tiny ImageNet}, which is a tiny version of ImageNet consisting of 3-channel color images with size of $64\times64$ belonging to 200 classes. Each class has 500 training images and 50 validation images. We use the comparison methods include {\em Undefended Model (UM)}, {\em adversarial training (AT)}~\cite{madry2017towards}, {\em adversarial logit pairing (ALP)}~\cite{kannan2018}, {\em single directional metric learning (SML)}~\cite{Mao2019}, {\em Bilateral}~\cite{Wang2019} and {\em feature scattering (FS)}~\cite{zhang2019defense}. To reduce the computational cost, we use ResNet-50 model as the same as SML. The learning rate $\gamma$ is initialized as 0.1, and decays at 30 epoch. We retrain and evaluate the models of bilateral and feature scatter using the existing codes.

From Table~\ref{tab:tiny_imagenet_result}, we can see that all methods show relatively poor performance on Tiny ImageNet. While our method outperforms others by a small margin, our performance can only achieve $\sim$20$\%$, which is not good enough for practical usage. It suggests that Tiny ImageNet is a difficult dataset due to its large class numbers and small sample size in each class. There is a large room to improve on this difficult dataset.

\begin{table*}[htbp]\small
\centering
\caption{\label{tab:tiny_imagenet_result} Evaluation results on Tiny ImageNet, under seven widely used attacks, as well as the results on clean images. The best accuracy for each attack is illustrated as bold. All attack budgets in training are $\epsilon=8$ by default for an apples to apples comparison.}
\begin{tabular}{c|cccccccc}
\hline \hline
\multicolumn{9}{c}{Tiny ImageNet} \\
\hline
 Attacks(steps) & clean & FGSM & BIM(10)  & PGD (10) & PGD (20) & CW (10) & CW (20) & MIM (40) \\
\hline 
UM & \textbf{64.62$\%$}  & 3.93$\%$  & 0.17$\%$  & 0.10$\%$  & 0.07$\%$  & 0$\%$  & 0$\%$  & 0.57$\%$ \\
Bilateral~\cite{Wang2019} & 58.70$\%$  & 30.81$\%$  & 20.98$\%$  & 19.73$\%$  & 18.98$\%$  & 15.19$\%$  & 14.61$\%$  & 22.47$\%$ \\
FS~\cite{zhang2019defense} & 53.81$\%$  & 30.06$\%$  & 20.59$\%$  & 19.46$\%$  & 18.52$\%$  & 15.53$\%$  & 14.68$\%$  & 22.40$\%$ \\
AT~\cite{madry2017towards} & 42.29$\%$  & 26.08$\%$  & 20.41$\%$  & 19.99$\%$  & 19.59$\%$  & 17.17$\%$  & 16.92$\%$  & 21.15$\%$  \\
ALP~\cite{kannan2018} & 41.53$\%$  & 21.53$\%$  & 20.03$\%$  & 20.18$\%$  & 19.96$\%$  & 16.80$\%$  & -  & 19.85$\%$ \\
SML~\cite{Mao2019} & 40.89$\%$  & 22.12$\%$  & 20.77$\%$  & 20.89$\%$  & 20.71$\%$  & 17.48$\%$  & -  & 20.69$\%$ \\
\hline
AGKD-BML & 53.21$\%$  & \textbf{31.39$\%$}  & \textbf{23.55$\%$}  & \textbf{22.68$\%$}  & \textbf{21.78$\%$}  & \textbf{18.8$\%$}  & \textbf{18.03$\%$}  & \textbf{24.71$\%$} \\
\hline \hline
\end{tabular}
\end{table*}

\section{Additional Results of AGKD-BML Model Against AutoAttack (AA)~\cite{AA_atack_2020}}
In Table~\ref{tab:aa}, we provide additional results of AGKD-BML model trained on 10-step attacks against AutoAttack (AA)~\cite{AA_atack_2020} which is an ensemble of four diverse attacks. We compare two Wide ResNet~\cite{ZagoruykoK16} structures, \ie, WRN-28-10 and WRN-34-10, as well as two different learning rate decay epochs, \ie, 100 and 150. For our AGKD-BML models trained with large-number-step attacks, we utilize the MART loss~\cite{MART_2020} which explicitly emphasizes misclassified examples. Following the suggestions in~\cite{rice2020overfitting, MART_2020} that the best performance is usually on a few epochs after the first learning rate decay, we stop our training at 5 epochs after the first learning rate decay. From Table~\ref{tab:aa}, we can see that with more layers, \ie 34 v.s. 28, the model usually performs better in terms of the accuracy against AA.
 
\begin{table}[htbp]
\small
\centering
\caption{\label{tab:aa} Evaluation results of AGKD-BML against AutoAttack (AA), on CIFAR-10 dataset, with different layers, learning rate and decay points. All results are evaluated by the models trained on 10-step attacks.}
\begin{tabular}{c|c|cc}
\hline \hline
\multicolumn{4}{c}{CIFAR-10} \\
\hline
 & Networks & Decay point & AA \\
\hline
 \multirow{4}{*}{AGKD-BML-10} & \multirow{2}{*}{WRN-28-10} & 100 & 50.80$\%$ \\
 &  & 150 & 50.73$\%$ \\
\cline{2-4}
 & \multirow{2}{*}{WRN-34-10} & 100 & 51.05$\%$ \\
 &  & 150 & 51.63$\%$ \\

\hline \hline
\end{tabular}
\end{table}

\section{Class-irrelevant attention distillation}

In our work, we use the class-\textit{irrelevant} attention to transfer the regions that the model focuses on, regardless of which class makes the contribution. By contrast, the class-\textit{relevant} attention map shows the attention region related to a specific class. 
An adversarial example (AE) fools a neural network (NN) by adding intentionally designed perturbations, which are further augmented by the NN to make the values of false class (actual prediction on AE) related attention map surpass that of original class, and thus make the NN misclassify the sample. 
We argue that transferring the information of class-relevant attention map is problematic: \\ 
1) {\it For the original class, transferring the class relevant attention} has limited effects since AE hurts much less the original class attention map than the false class one. More importantly, it rarely reduces the dominated responses of the false class related maps which limits the effects of correcting the misclassification. \\
2) {\it For the false class, transferring the class relevant attention} enforces the false class attention map to focus on the regions where objects of the original class locate, and it does not modify the attention regions of the original class.

We conducted the experiment to compare the class relevant/irrelevant attention distillation. We trained models by the class-relevant attention maps generated by Grad-CAM corresponding to both original and false classes. In addition to our AGKD, we also trained a model by another class irrelevant attention map generated by averaging all CAM maps of all classes~\cite{zhou2016learning}. The results in Table~\ref{tab:cls_ir} demonstrate better performance achieved by the class-irrelevant attention distillation. We chose AGKD over CAM-avg since CAM-avg is computationally heavy.

\begin{table}[htbp]
\small
\centering
\caption{\label{tab:cls_ir} Comparisons of class-relevant and -irrelevant attention.}
\begin{tabular}{cc|cc}
\hline \hline
\multicolumn{2}{c|}{Class-relevant} & \multicolumn{2}{c}{Class-irrelevant} \\
\hline
GC-orig & GC-false & CAM-avg & AGKD \\
\hline
59.06$\%$  & 58.86$\%$  &  66.55$\%$   & 65.93$\%$ \\

\hline \hline
\end{tabular}
\end{table}

\section{Comparison of Running Time}
We provide the training time of bilateral~\cite{Wang2019}, feature scatter~\cite{zhang2019defense}, AT~\cite{madry2017towards}, SML~\cite{Mao2019} and our AGKD-BML model on CIFAR-10 dataset. In Table~\ref{tab:running_time}, we provide implementation platforms, training time (seconds) per epoch, number of epochs for training, total time (hours) for training, as well as the number of the steps to get the adversarial examples for each model. All running times are evaluated on one Nvidia V100 GPU with 32GB memory. Once trained, testing times for all the models are approximately the same, although it shows in Table~\ref{tab:running_time} that our model takes more time in training. In the security applications, the training time is not critical compared to the accuracy. Therefore, 1.2 days of training time of AGKD-BML model is acceptable.

Furthermore, we also evaluated the performance of different models with the same running time. FS and Bilateral originally use only one-step attacks. Therefore, we trained FS with 2-step attack (470s) and Bilateral with 4-step attack (453s), and got accuracy of 70.51$\%$ and 59.99$\%$ (compared to ours: \textbf{71.02$\%$}, 528s), respectively.

\begin{table*}[htbp]
\small
\centering
\caption{\label{tab:running_time} Training time comparison.}
\begin{tabular}{c|ccccc}
\hline \hline
\multicolumn{6}{c}{CIFAR-10} \\
\hline
 & $\sharp$ of steps  & platform & seconds / epoch  & $\sharp$ of epochs & total (hours)\\
\hline 
Bilateral~\cite{Wang2019} & 1 & TensorFlow & 211s & 200 & 11.7h\\
FS~\cite{zhang2019defense} & 1 & PyTorch & 342s & 200 & 19.0h\\
AT~\cite{madry2017towards} & 7  & PyTorch &  502s & 200 & 27.9h\\
SML~\cite{Mao2019} & 7  & TensorFlow &  2234s  & 55 & 34.1h\\
\hline
AGKD-BML & 2  & PyTorch &  528s & 200 & 29.3h\\
\hline \hline
\end{tabular}
\end{table*}

\section{Comparison of A Latest Model}
In~\cite{cheng2020cat}, the authors proposed a customized adversarial training (CAT) model, which adaptively tunes a suitable $\epsilon$ for each sample during the adversarial training procedure. However, the authors do not provide systematical results of the same experiment settings as that in~\cite{Mao2019,Wang2019, zhang2019defense}, instead, they only provide the results under PGD and CW attacks on CIFAR-10 dataset. Therefore, we report our results of the same experimental setting as CAT in Table~\ref{tab:CAT}. CAT has two variants, 1) ``CAT-CE" applies standard cross entropy loss as used in traditional adversarial training models~\cite{madry2017towards}, and 2) ``CAT-MIX" applies both cross entropy loss and CW loss~\cite{carlini2017towards}. Note that CAT used an adaptive $\epsilon$ for training, while our results are given with a fixed $\epsilon=4$. From the table, we can see that our proposed AGKD-BML model consistently outperformed ``CAT-CE", and ``CAT-MIX" except under ``CW" attack. This is because both AGKD-BML and CAT-CE only apply cross entropy which is used in PGD attack, and ``CAT-MIX" includes both cross entropy loss and CW loss that is used in CW attack.

\begin{table*}[htbp]
\small
\caption{\label{tab:CAT} Comparing with customized adversarial training (CAT). Note that ``CAT-MIX" applies CW as part of its loss.}
\centering
\begin{tabular}{c|ccc|cc}
\hline \hline
\multicolumn{6}{c}{CIFAR-10} \\
\hline
\multirow{2}{*}{Models}&\multicolumn{3}{c|}{White-box} & \multicolumn{2}{c}{Black-box} \\
\cline{2-6}
& clean & PGD  & CW & VGG-16 & Wide ResNet\\
\hline 
CAT-CE~\cite{cheng2020cat} (adaptive $\epsilon$) & 93.48$\%$ & 73.38$\%$  & 61.88$\%$  & 86.58$\%$ & 88.66$\%$\\
CAT-MIX~\cite{cheng2020cat} (adaptive $\epsilon$) & 89.61$\%$ & 73.16$\%$  & \textbf{71.67$\%$}  & - & - \\
AGKD-BML (fixed $\epsilon=4$) & \textbf{95.04$\%$} & \textbf{77.45$\%$} & 69.06$\%$  & \textbf{89.12$\%$}  & \textbf{91.98$\%$} \\
\hline \hline
\end{tabular}
\end{table*}

\section{Evaluation of k-Nearest Neighbor (k-NN) classifier}\label{sec:knn}

\begin{table*}[htbp]
\small
\caption{\label{tab:knn} Evaluation and comparison between k-NN and softmax classifiers (k-NN/softmax). Our proposed AGKD-BML with k-NN classifier consistently outperforms other comparison methods, and shows very similar accuracy with softmax classifier.}
\centering
\begin{tabular}{c|cc|cc|cc}
\hline \hline
\multirow{2}{*}{Models}&\multicolumn{2}{c|}{CIFAR-10} & \multicolumn{2}{c|}{SVHN} & \multicolumn{2}{c}{Tiny ImageNet}\\
\cline{2-7}
& clean & PGD (20)  & clean & PGD (20)  & clean & PGD (20)  \\
\hline 
AT~\cite{madry2017towards} & 87.1$\%$ / 86.2$\%$ & 47.5$\%$ / 45.6$\%$ & 91.5$\%$ / 91.6$\%$  & 45.8$\%$ / 45.6$\%$ & 36.6$\%$ / 42.3$\%$ & 20.2$\%$ / 19.6$\%$\\
ALP~\cite{kannan2018} & 89.6$\%$ / 89.8$\%$ & 48.9$\%$ / 48.5$\%$  & 91.4$\%$ / 91.3$\%$ & 52.0$\%$ / 52.2$\%$ & 35.2$\%$ / 41.5$\%$ & 20.3$\%$ / 20.0$\%$\\
SML~\cite{Mao2019} & 86.3$\%$ / 86.2$\%$ & 51.7$\%$ / 51.6$\%$ & 84.3$\%$ / 84.0$\%$ & 52.0$\%$ / 51.9$\%$ & 34.0$\%$ / 40.6$\%$& 20.7$\%$ / 20.7$\%$\\
AGKD-BML & \textbf{91.9$\%$} / \textbf{92.0$\%$} & \textbf{71.1$\%$} / \textbf{71.0$\%$} & \textbf{95.1$\%$}  / \textbf{95.0$\%$}   & \textbf{75.1$\%$} / \textbf{74.9$\%$} & \textbf{51.8$\%$} / \textbf{53.2$\%$} & \textbf{21.0$\%$} / \textbf{22.7$\%$}\\
\hline \hline
\end{tabular}
\end{table*}

We conduct the experiments that apply k-Nearest Neighbor (k-NN) method as the classifier following~\cite{Mao2019}. 
We utilize the feature vectors from the penultimate layer to perform the k-NN classifier for all the models with $k=50$. 

In Table~\ref{tab:knn}, we show the k-NN classifier accuracy, as well as corresponding softmax accuracy, between four models, \ie, AT~\cite{madry2017towards} ALP~\cite{kannan2018} SML~\cite{Mao2019} and proposed AGKD-BML, on CIFAR-10, SVHN and Tiny ImageNet datasets. Our model consistently achieves higher accuracy on all three datasets. Moreover, the k-NN classifier usually performs very similarly as softmax. These quantitative results, coupled with the visualization illustrations in next section (Section~\ref{sec:supp_tsne}), demonstrate that AGKD-BML is able to obtain better representation in the latent feature space than other comparison methods, and the accuracy does benefit from the good representation, rather than the classifier.


\section{Visualization Analysis with t-SNE}\label{sec:supp_tsne}

\begin{figure*}[!t]
	\centering
	\begin{tabular}{@{\hspace{.0mm}}c@{\hspace{1.75mm}} @{\hspace{.0mm}}c@{\hspace{.0mm}} @{\hspace{.0mm}}c@{\hspace{.0mm}}}
	    \vspace{1mm} \hspace{3.1cm} UM \hspace{3cm} SML \hspace{2.5cm} BML \hspace{2.2cm} AGKD-BML \\
	    \includegraphics[width=\linewidth]{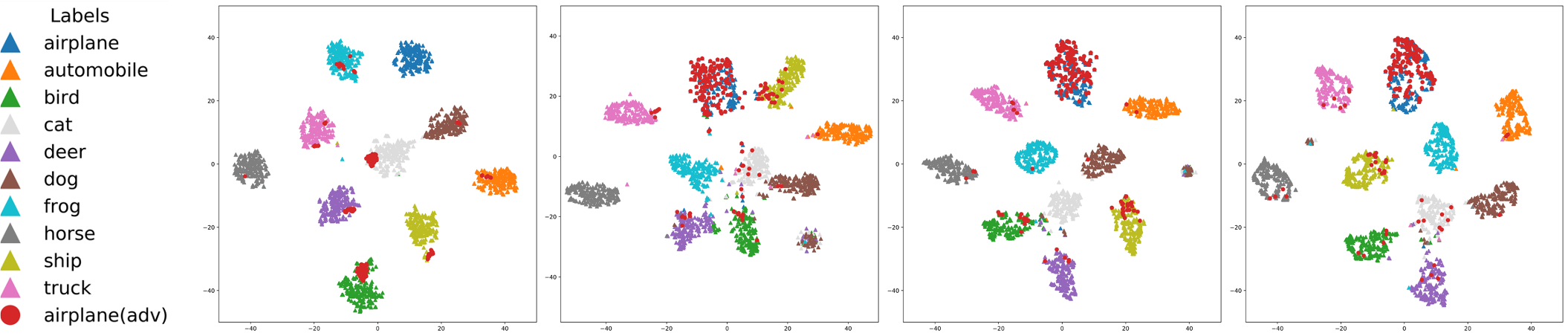} \\
		\includegraphics[width=\linewidth]{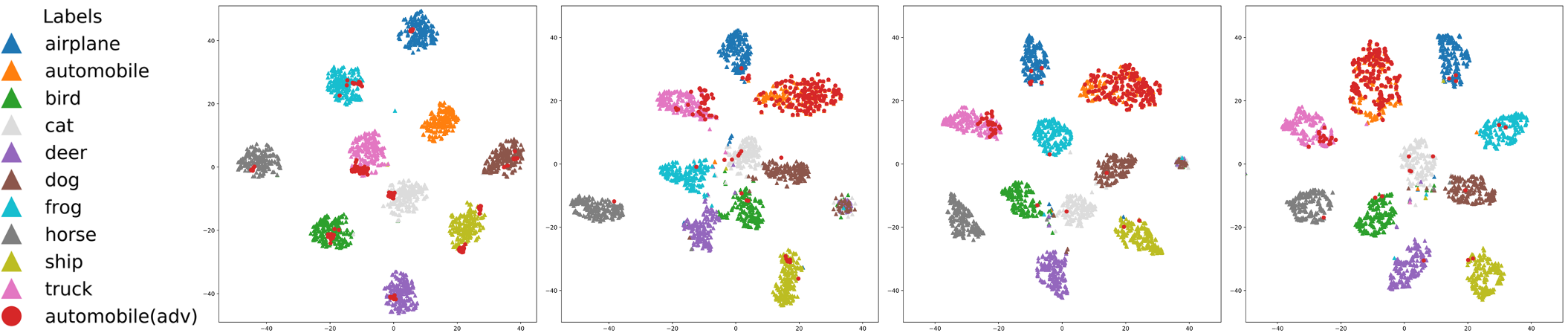} \\
		\includegraphics[width=\linewidth]{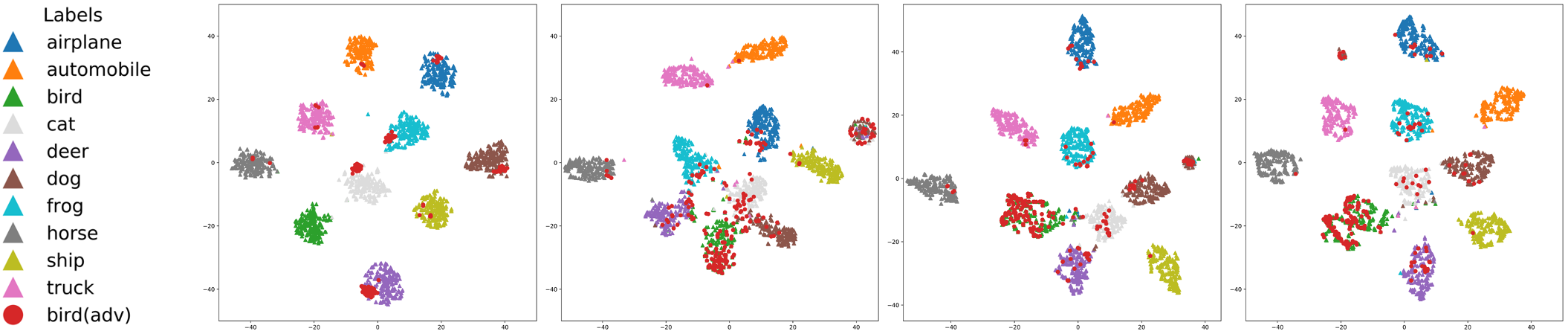} \\
		\includegraphics[width=\linewidth]{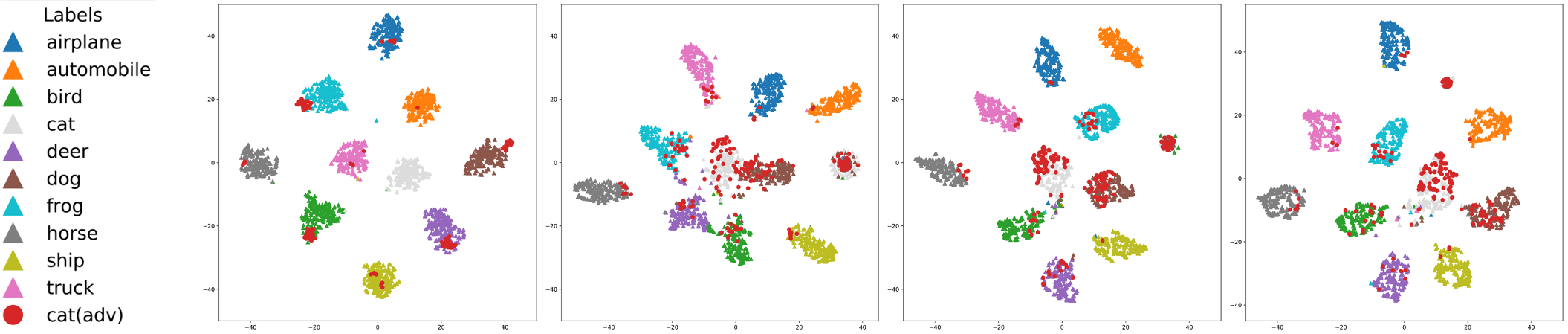} \\
		\includegraphics[width=\linewidth]{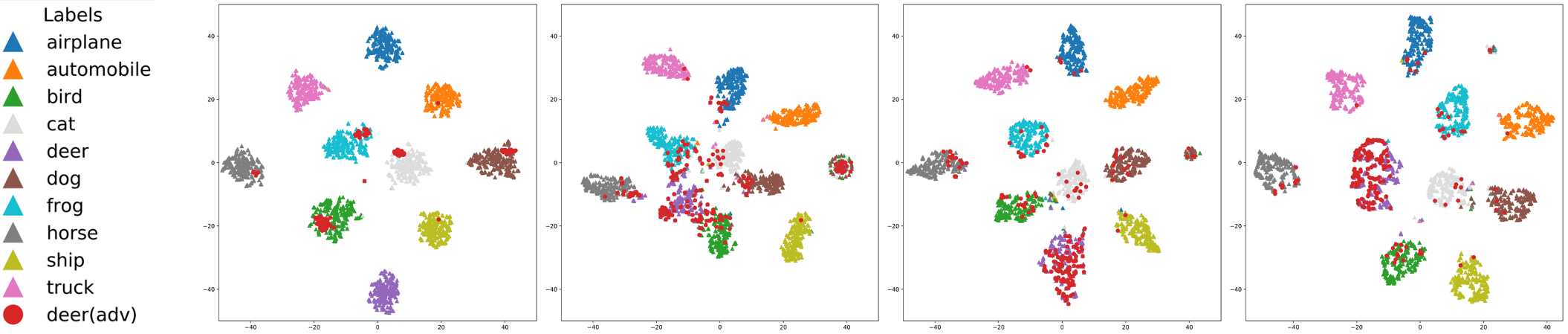} \\
	\end{tabular}
	\caption{t-SNE plots for illustrating the sample representations in feature space. The adversarial example is airplane, automobile, bird, cat and deer in each row, respectively, under PGD-20 attack.}.
	\label{fig:PGD-20-0_4}
\end{figure*}

\begin{figure*}[!t]
	\centering
	\begin{tabular}{@{\hspace{.0mm}}c@{\hspace{1.75mm}} @{\hspace{.0mm}}c@{\hspace{.0mm}} @{\hspace{.0mm}}c@{\hspace{.0mm}}}
	    \vspace{1mm} \hspace{3.1cm} UM \hspace{3cm} SML \hspace{2.5cm} BML \hspace{2.2cm} AGKD-BML \\
	    \includegraphics[width=\linewidth]{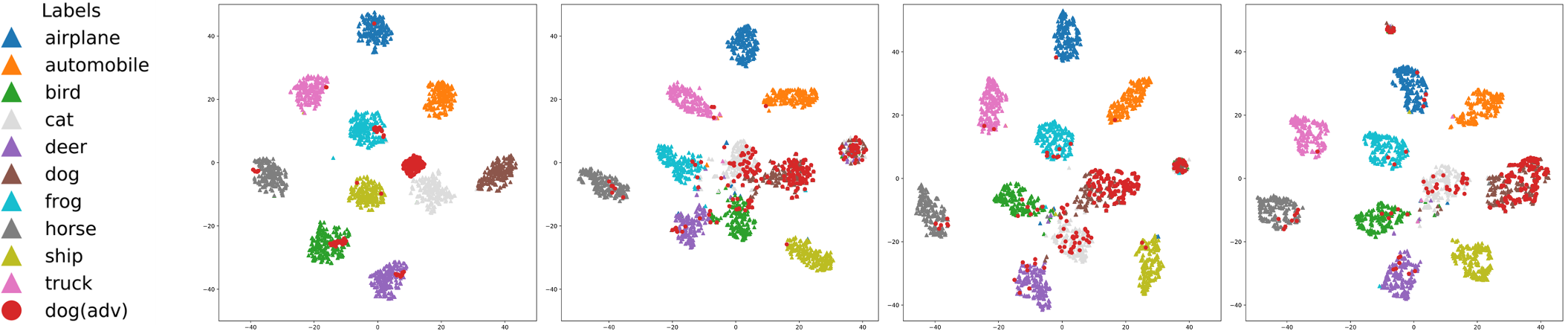} \\
		\includegraphics[width=\linewidth]{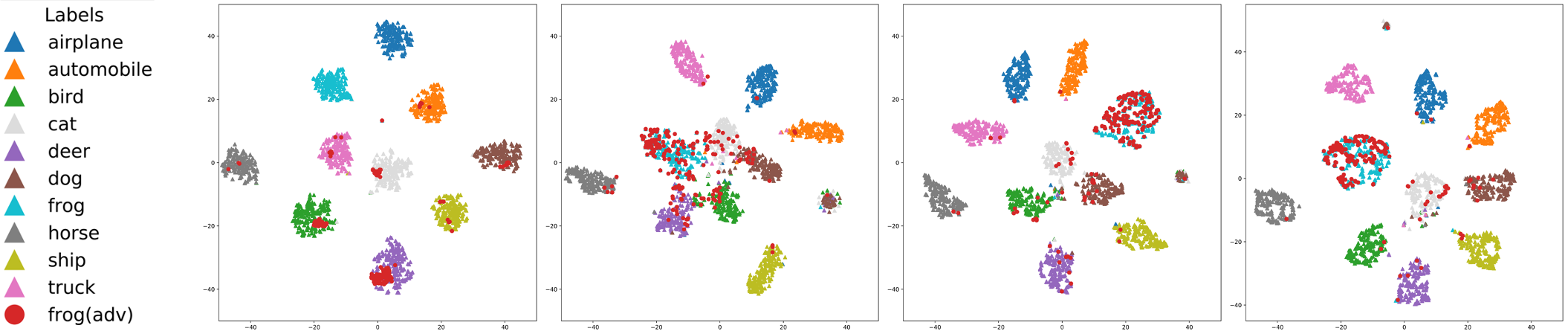} \\
		\includegraphics[width=\linewidth]{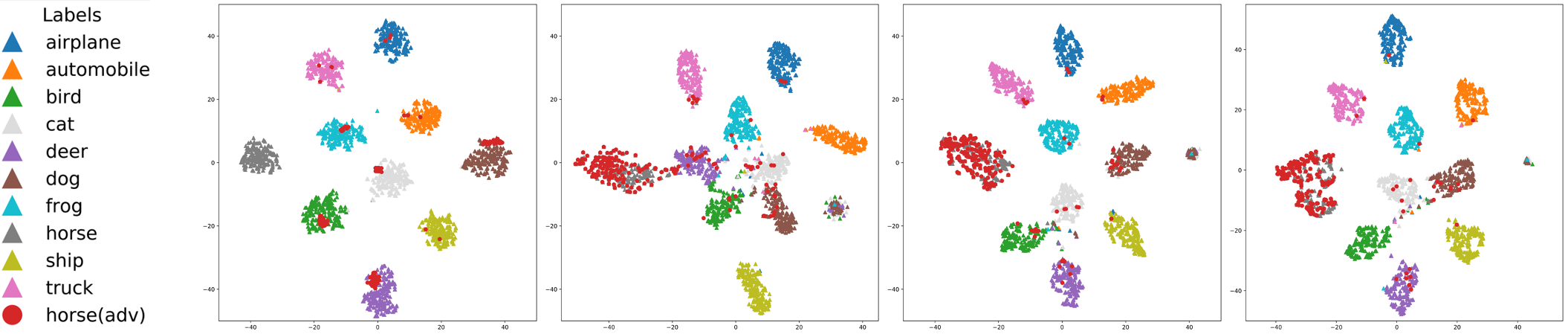} \\
		\includegraphics[width=\linewidth]{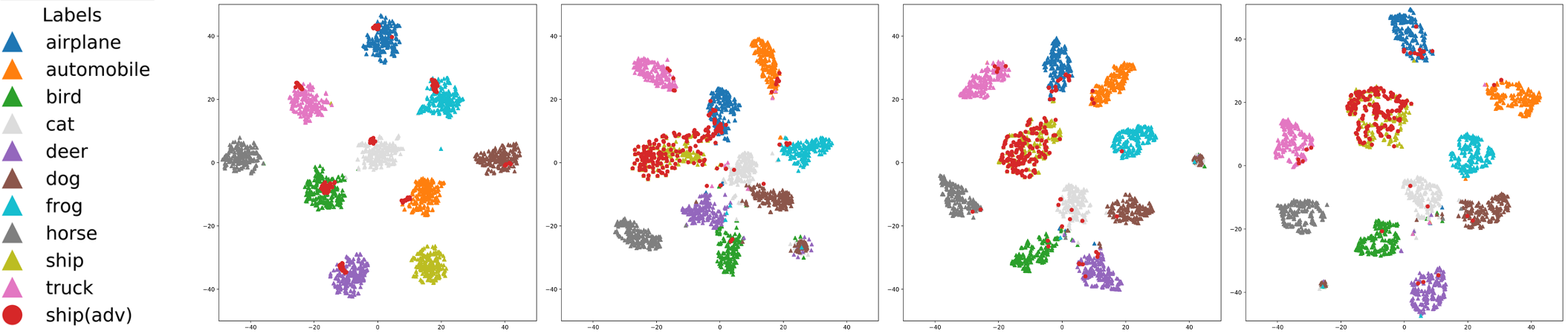} \\
		\includegraphics[width=\linewidth]{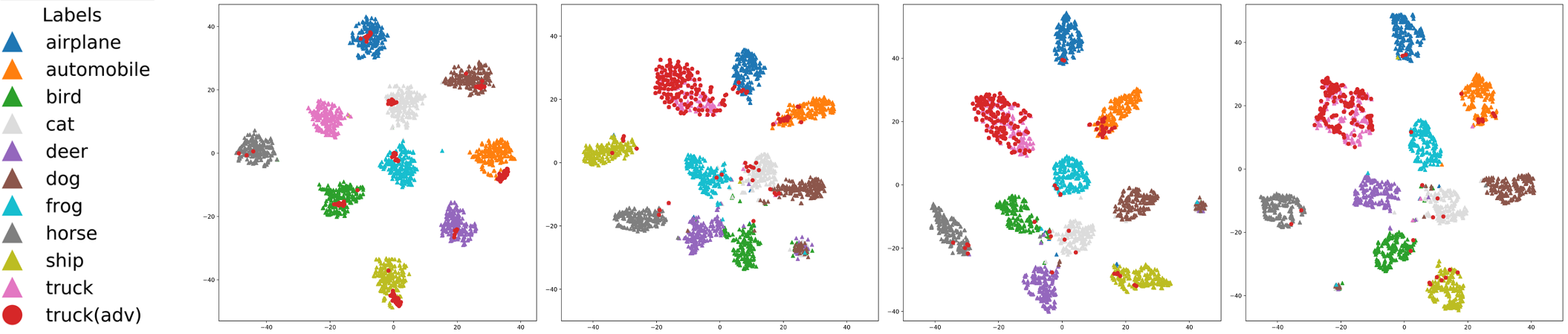} \\
	\end{tabular}
	\caption{t-SNE plots for illustrating the sample representations in feature space. The adversarial example is dog, frog, horse, ship and truck in each row, respectively, under PGD-20 attack.}.
	\label{fig:PGD-20-5_9}
\end{figure*}

\begin{figure*}[!t]
	\centering
	\begin{tabular}{@{\hspace{.0mm}}c@{\hspace{1.75mm}} @{\hspace{.0mm}}c@{\hspace{.0mm}} @{\hspace{.0mm}}c@{\hspace{.0mm}}}
	    \vspace{1mm} \hspace{3.1cm} UM \hspace{3cm} SML \hspace{2.5cm} BML \hspace{2.2cm} AGKD-BML \\
	    \includegraphics[width=\linewidth]{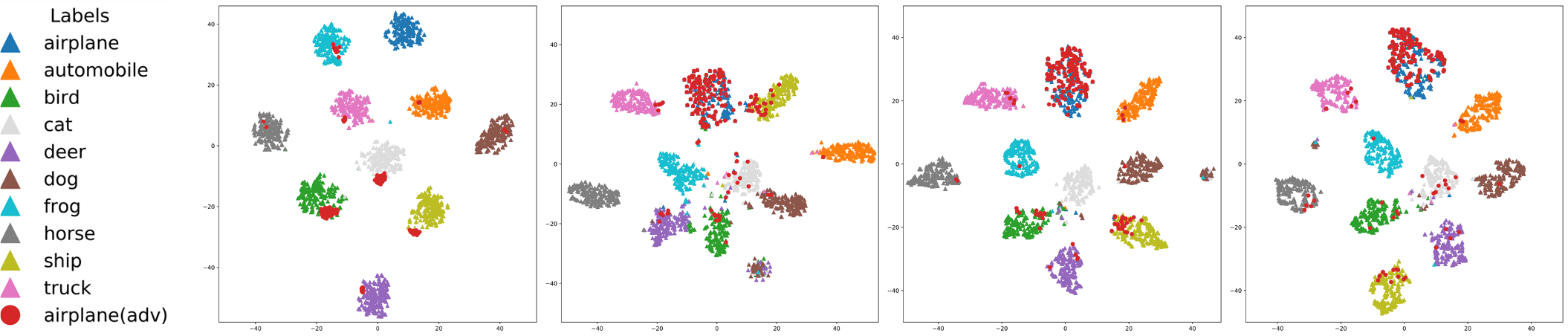} \\
		\includegraphics[width=\linewidth]{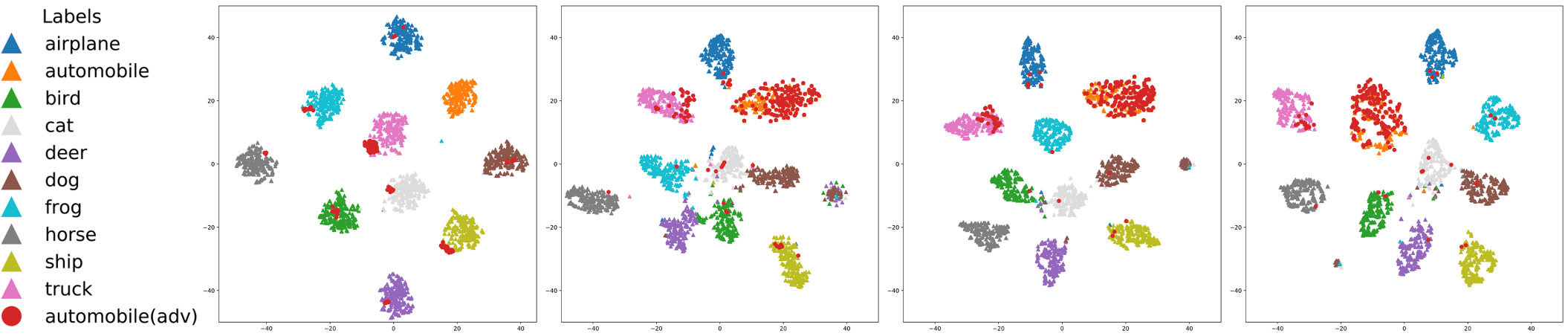} \\
		\includegraphics[width=\linewidth]{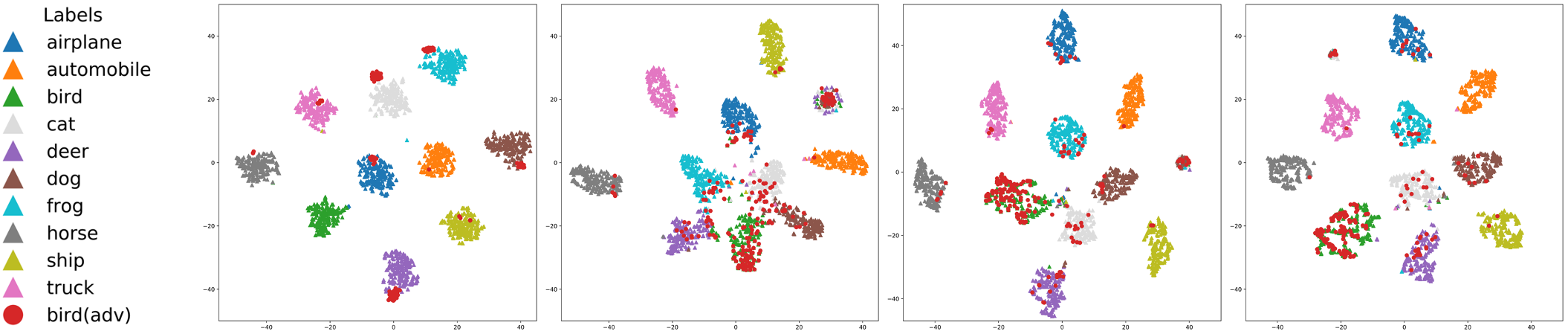} \\
		\includegraphics[width=\linewidth]{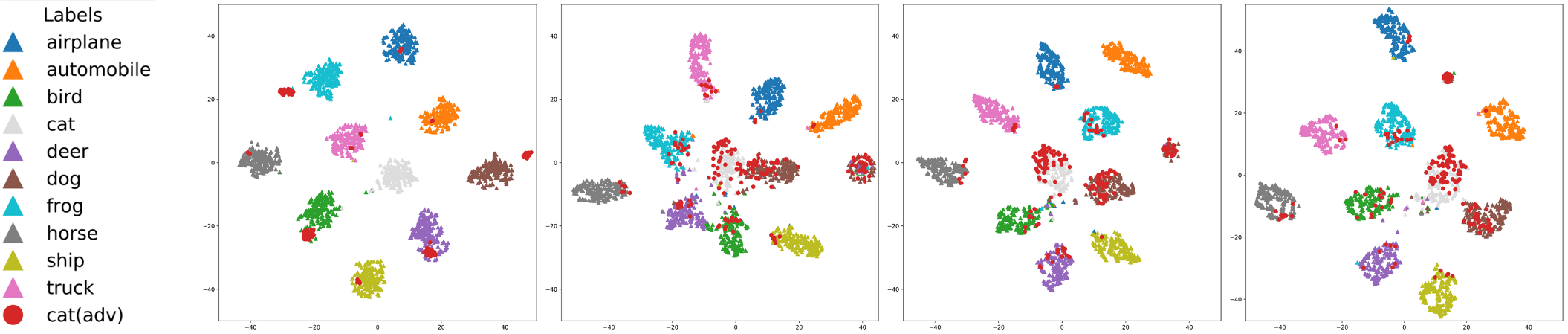} \\
		\includegraphics[width=\linewidth]{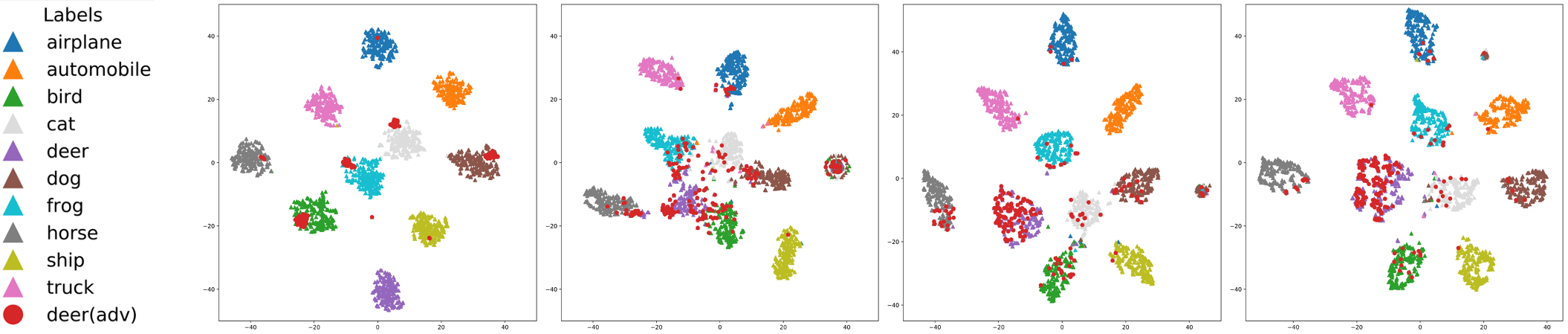} \\
		
	\end{tabular}
	\caption{t-SNE plots for illustrating the sample representations in feature space. The adversarial example is airplane, automobile, bird, cat and deer in each row, respectively, under PGD-100 attack.}.
	\label{fig:PGD-100-0_4}
\end{figure*}

\begin{figure*}[!t]
	\centering
	\begin{tabular}{@{\hspace{.0mm}}c@{\hspace{1.75mm}} @{\hspace{.0mm}}c@{\hspace{.0mm}} @{\hspace{.0mm}}c@{\hspace{.0mm}}}
	    \vspace{1mm} \hspace{3.1cm} UM \hspace{3cm} SML \hspace{2.5cm} BML \hspace{2.2cm} AGKD-BML \\
	    \includegraphics[width=\linewidth]{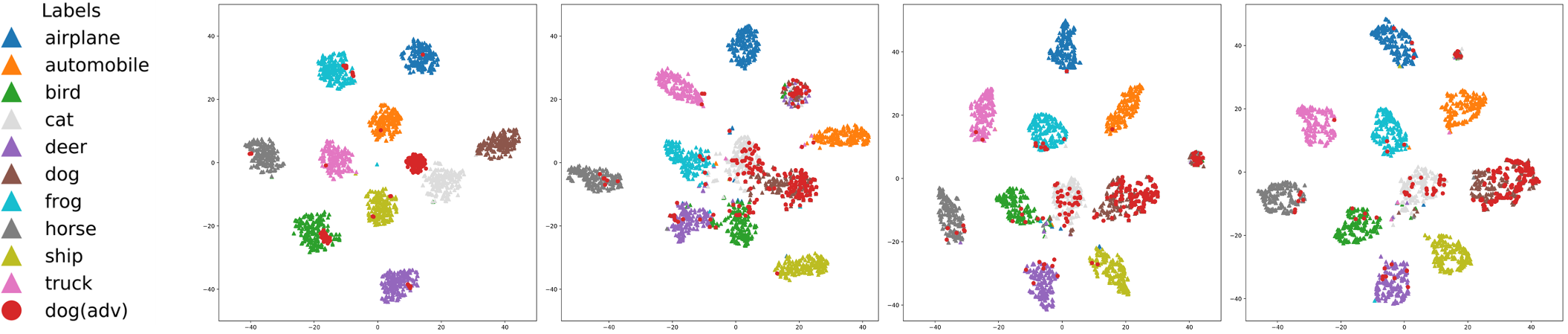} \\
		\includegraphics[width=\linewidth]{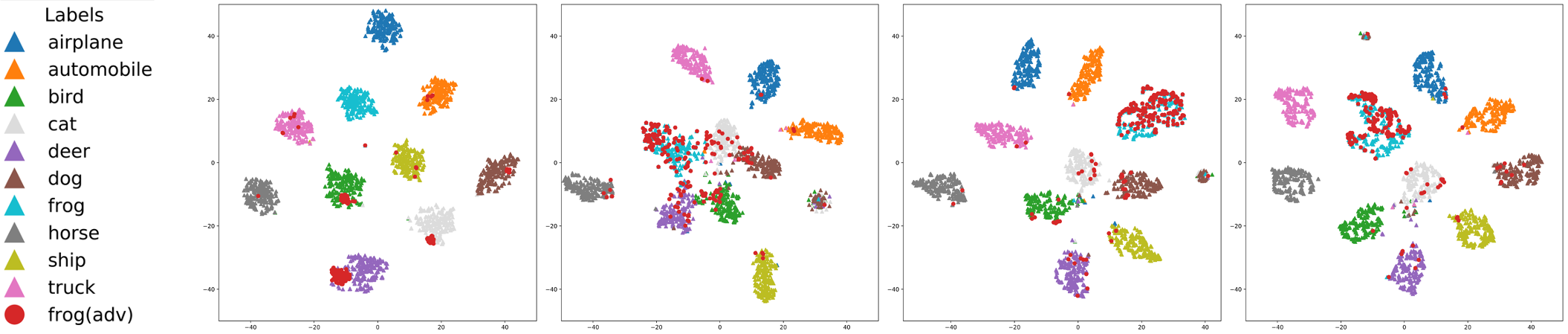} \\
		\includegraphics[width=\linewidth]{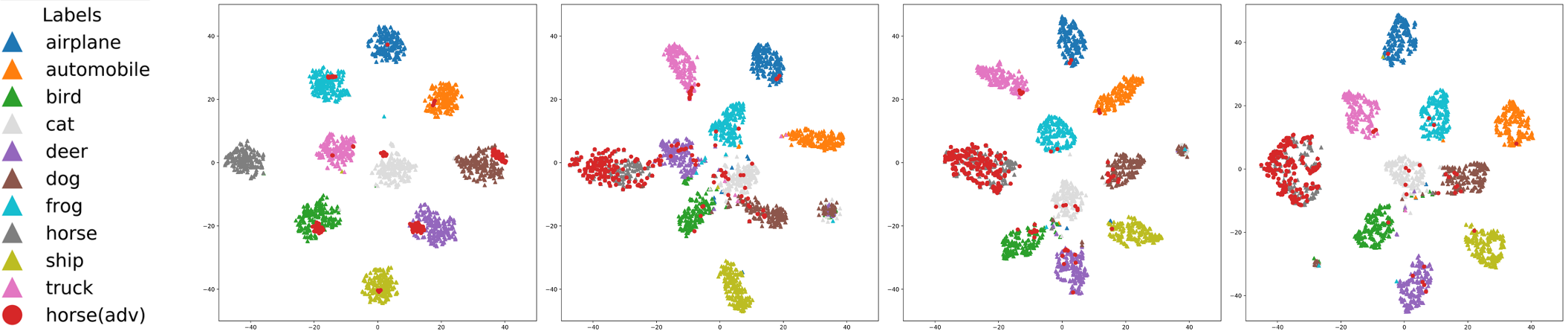} \\
		\includegraphics[width=\linewidth]{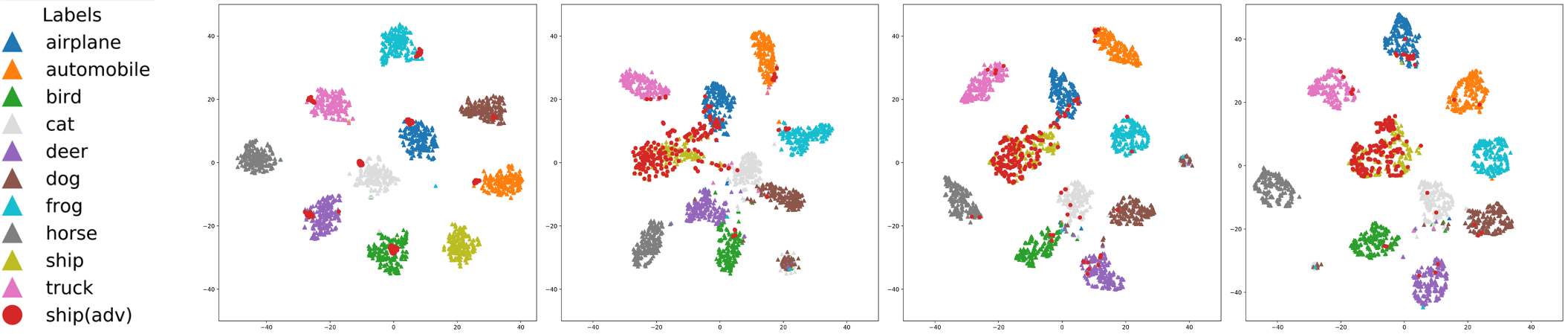} \\
		\includegraphics[width=\linewidth]{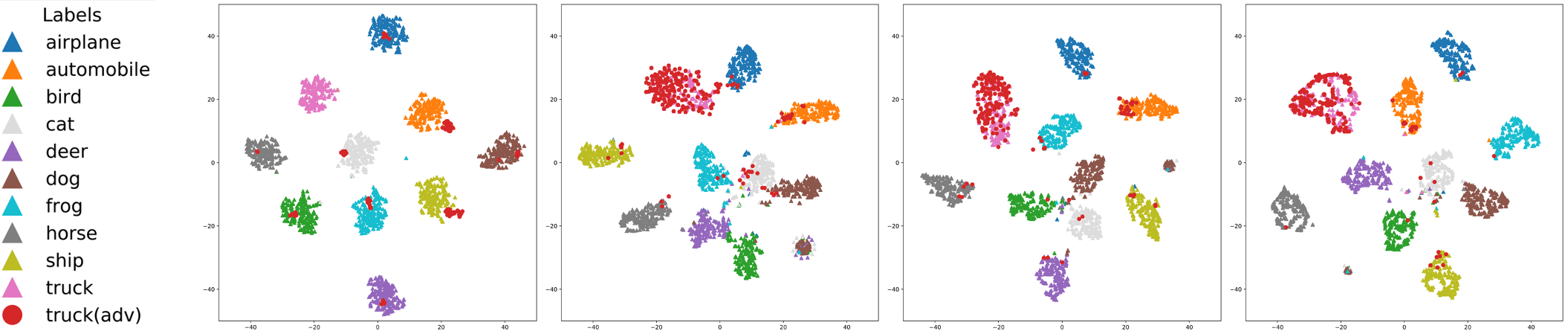} \\
	\end{tabular}
	\caption{t-SNE plots for illustrating the sample representations in feature space. The adversarial example is dog, frog, horse, ship and truck in each row, respectively, under PGD-100 attack.}.
	\label{fig:PGD-100-5_9}
\end{figure*}

In Figure~\ref{fig:PGD-20-0_4},~\ref{fig:PGD-20-5_9},~\ref{fig:PGD-100-0_4},~\ref{fig:PGD-100-5_9}, we provide the t-SNE plots to show the sample representations in feature space for all attacked classes under PGD-20 and PGD-100 attacks. The triangle points with different colors represent the clean images in different classes, while the red circle points are the adversarial examples under attack. We show the adversarial examples of airplane, automobile, bird, cat and deer under PGD-20 and PGD-100 attack in Figure~\ref{fig:PGD-20-0_4} and Figure~\ref{fig:PGD-100-0_4}, respectively, and the adversarial examples of dog, frog, horse, ship and truck under PGD-20 and PGD-100 attack in Figure~\ref{fig:PGD-20-5_9} and Figure~\ref{fig:PGD-100-5_9}, respectively.

From these figures, the same observations can be seen as in the main text, which we would like to emphasize as following: 
\begin{itemize}
\item In the first column of all the figures, ``UM" is almost {\em non-robust} to the adversarial examples, as it shows that all the adversarial examples are far away from their original class, and fit into the distributions of other classes. 

\item In the second and third columns of all the figures, while SML does pull many of the adversarial examples back to their original class,  BML keeps better separations between different classes, and has much less amount of adversarial examples located far away compared to SML. 

\item By integrating both AGKD and BML, AGKD-BML pulls most of the adversarial examples back to their original class, while keeps best separation between classes overall, as shown in the fourth column of all the figures. 
\end{itemize}

{\small
\bibliographystyle{ieee_fullname}
\bibliography{ICCV}

\begin{thebibliography}{10}\itemsep=-1pt

\bibitem{andriushchenko2020square}
Maksym Andriushchenko, Francesco Croce, Nicolas Flammarion, and Matthias Hein.
\newblock Square attack: a query-efficient black-box adversarial attack via
  random search.
\newblock In {\em ECCV}, pages 484--501. Springer, 2020.

\bibitem{athalye2018obfuscated}
Anish Athalye, Nicholas Carlini, and David Wagner.
\newblock Obfuscated gradients give a false sense of security: Circumventing
  defenses to adversarial examples.
\newblock In {\em ICML}, pages 274--283, 2018.

\bibitem{CAS_2021}
Yang Bai, Yuyuan Zeng, Yong Jiang, Shu-Tao Xia, Xingjun Ma, and Yisen Wang.
\newblock Improving adversarial robustness via channel-wise activation
  suppressing.
\newblock In {\em ICLR}, 2021.

\bibitem{carlini2017towards}
Nicholas Carlini and David Wagner.
\newblock Towards evaluating the robustness of neural networks.
\newblock In {\em 2017 IEEE symposium on security and privacy (SP)}, pages
  39--57, 2017.

\bibitem{carmon2019unlabeled}
Yair Carmon, Aditi Raghunathan, Ludwig Schmidt, Percy Liang, and John~C Duchi.
\newblock Unlabeled data improves adversarial robustness.
\newblock {\em NeurIPS}, 2019.

\bibitem{cheng2020cat}
Minhao Cheng, Qi Lei, Pin-Yu Chen, Inderjit Dhillon, and Cho-Jui Hsieh.
\newblock Cat: Customized adversarial training for improved robustness.
\newblock {\em arXiv:2002.06789}, 2020.

\bibitem{cohen2019certified}
Jeremy Cohen, Elan Rosenfeld, and Zico Kolter.
\newblock Certified adversarial robustness via randomized smoothing.
\newblock In {\em ICML}, pages 1310--1320, 2019.

\bibitem{collobert2008unified}
Ronan Collobert and Jason Weston.
\newblock A unified architecture for natural language processing: Deep neural
  networks with multitask learning.
\newblock In {\em ICML}, pages 160--167, 2008.

\bibitem{croce2020minimally}
Francesco Croce and Matthias Hein.
\newblock Minimally distorted adversarial examples with a fast adaptive
  boundary attack.
\newblock In {\em ICML}, pages 2196--2205, 2020.

\bibitem{AA_atack_2020}
Francesco Croce and Matthias Hein.
\newblock Reliable evaluation of adversarial robustness with an ensemble of
  diverse parameter-free attacks.
\newblock In {\em ICML}, pages 2206--2216, 2020.

\bibitem{ding2020mma}
Gavin~Weiguang Ding, Yash Sharma, Kry Yik~Chau Lui, and Ruitong Huang.
\newblock Mma training: Direct input space margin maximization through
  adversarial training.
\newblock In {\em ICLR}, 2020.

\bibitem{dong2018boosting}
Yinpeng Dong, Fangzhou Liao, Tianyu Pang, Hang Su, Jun Zhu, Xiaolin Hu, and
  Jianguo Li.
\newblock Boosting adversarial attacks with momentum.
\newblock In {\em CVPR}, pages 9185--9193, 2018.

\bibitem{eykholt2018robust}
Kevin Eykholt, Ivan Evtimov, Earlence Fernandes, Bo Li, Amir Rahmati, Chaowei
  Xiao, Atul Prakash, Tadayoshi Kohno, and Dawn Song.
\newblock Robust physical-world attacks on deep learning visual classification.
\newblock In {\em CVPR}, pages 1625--1634, 2018.

\bibitem{fawzi2015manitest}
Alhussein Fawzi and Pascal Frossard.
\newblock Manitest: Are classifiers really invariant?
\newblock In {\em BMVC}, pages 106.1--106.13, 2015.

\bibitem{goodfellow2009measuring}
Ian Goodfellow, Honglak Lee, Quoc Le, Andrew Saxe, and Andrew Ng.
\newblock Measuring invariances in deep networks.
\newblock In {\em NeurIPS}, pages 646--654, 2009.

\bibitem{goodfellow2014explaining}
Ian Goodfellow, Jonathon Shlens, and Christian Szegedy.
\newblock Explaining and harnessing adversarial examples.
\newblock In {\em ICLR}, 2015.

\bibitem{hinton2012deep}
Geoffrey Hinton, Li Deng, Dong Yu, George~E Dahl, Abdel-rahman Mohamed, Navdeep
  Jaitly, Andrew Senior, Vincent Vanhoucke, Patrick Nguyen, Tara~N Sainath,
  et~al.
\newblock Deep neural networks for acoustic modeling in speech recognition: The
  shared views of four research groups.
\newblock {\em IEEE Signal processing magazine}, 29(6):82--97, 2012.

\bibitem{hinton2015distilling}
Geoffrey Hinton, Oriol Vinyals, and Jeff Dean.
\newblock Distilling the knowledge in a neural network.
\newblock {\em arXiv:1503.02531}, 2015.

\bibitem{huang2020universal}
Lifeng Huang, Chengying Gao, Yuyin Zhou, Cihang Xie, Alan~L Yuille, Changqing
  Zou, and Ning Liu.
\newblock Universal physical camouflage attacks on object detectors.
\newblock In {\em CVPR}, pages 720--729, 2020.

\bibitem{kanbak2018geometric}
Can Kanbak, Seyed-Mohsen Moosavi-Dezfooli, and Pascal Frossard.
\newblock Geometric robustness of deep networks: analysis and improvement.
\newblock In {\em CVPR}, pages 4441--4449, 2018.

\bibitem{kannan2018}
Harini Kannan, Alexey Kurakin, and Ian Goodfellow.
\newblock Adversarial logit pairing.
\newblock {\em arXiv:1803.06373}, 2018.

\bibitem{kannan2018adversarial}
Harini Kannan, Alexey Kurakin, and Ian~J. Goodfellow.
\newblock Adversarial logit pairing.
\newblock {\em arXiv:1803.06373}, 2018.

\bibitem{NIPS2012_c399862d}
Alex Krizhevsky, Ilya Sutskever, and Geoffrey~E Hinton.
\newblock Imagenet classification with deep convolutional neural networks.
\newblock In {\em NeurIPS}, pages 1097--1105, 2012.

\bibitem{kurakin2016adversarial}
Alexey Kurakin, Ian Goodfellow, and Samy Bengio.
\newblock Adversarial examples in the physical world.
\newblock {\em ICLR Workshop}, 2017.

\bibitem{li2019improving}
Pengcheng Li, Jinfeng Yi, Bowen Zhou, and Lijun Zhang.
\newblock Improving the robustness of deep neural networks via adversarial
  training with triplet loss.
\newblock {\em arXiv:1905.11713}, 2019.

\bibitem{PASM21}
Ping Liu, Yuewei Lin, Zibo Meng, Lu Lu, Weihong Deng, Joey~Tianyi Zhou, and Yi
  Yang.
\newblock Point adversarial self-mining: A simple method for facial expression
  recognition.
\newblock {\em IEEE Transactions on Cybernetics}, pages 1--12, 2021.

\bibitem{lu2017safetynet}
Jiajun Lu, Theerasit Issaranon, and David Forsyth.
\newblock Safetynet: Detecting and rejecting adversarial examples robustly.
\newblock In {\em ICCV}, pages 446--454, 2017.

\bibitem{madry2017towards}
Aleksander Madry, Aleksandar Makelov, Ludwig Schmidt, Dimitris Tsipras, and
  Adrian Vladu.
\newblock Towards deep learning models resistant to adversarial attacks.
\newblock In {\em ICLR}, 2018.

\bibitem{Mao2019}
Chengzhi Mao, Ziyuan Zhong, Junfeng Yang, Carl Vondrick, and Baishakhi Ray.
\newblock Metric learning for adversarial robustness.
\newblock In {\em NeurIPS}, pages 480--491, 2019.

\bibitem{moosavi2016deepfool}
Seyed-Mohsen Moosavi-Dezfooli, Alhussein Fawzi, and Pascal Frossard.
\newblock Deepfool: a simple and accurate method to fool deep neural networks.
\newblock In {\em CVPR}, pages 2574--2582, 2016.

\bibitem{pan2020adversarial}
Pingbo Pan, Ping Liu, Yan Yan, Tianbao Yang, and Yi Yang.
\newblock Adversarial localized energy network for structured prediction.
\newblock In {\em Proceedings of the AAAI Conference on Artificial
  Intelligence}, volume~34, pages 5347--5354, 2020.

\bibitem{papernot2016limitations}
Nicolas Papernot, Patrick McDaniel, Somesh Jha, Matt Fredrikson, Z~Berkay
  Celik, and Ananthram Swami.
\newblock The limitations of deep learning in adversarial settings.
\newblock In {\em IEEE European symposium on security and privacy}, pages
  372--387, 2016.

\bibitem{rakin2019bit}
Adnan~Siraj Rakin, Zhezhi He, and Deliang Fan.
\newblock Bit-flip attack: Crushing neural network with progressive bit search.
\newblock In {\em ICCV}, pages 1211--1220, 2019.

\bibitem{rice2020overfitting}
Leslie Rice, Eric Wong, and Zico Kolter.
\newblock Overfitting in adversarially robust deep learning.
\newblock In {\em ICML}, pages 8093--8104. PMLR, 2020.

\bibitem{selvaraju2017grad}
Ramprasaath~R Selvaraju, Michael Cogswell, Abhishek Das, Ramakrishna Vedantam,
  Devi Parikh, and Dhruv Batra.
\newblock Grad-cam: Visual explanations from deep networks via gradient-based
  localization.
\newblock In {\em ICCV}, pages 618--626, 2017.

\bibitem{shafahi2019adversarial}
Ali Shafahi, Mahyar Najibi, Mohammad~Amin Ghiasi, Zheng Xu, John Dickerson,
  Christoph Studer, Larry~S Davis, Gavin Taylor, and Tom Goldstein.
\newblock Adversarial training for free!
\newblock In {\em NeurIPS}, pages 3358--3369, 2019.

\bibitem{song2018pixeldefend}
Yang Song, Taesup Kim, Sebastian Nowozin, Stefano Ermon, and Nate Kushman.
\newblock Pixeldefend: Leveraging generative models to understand and defend
  against adversarial examples.
\newblock In {\em ICLR}, 2018.

\bibitem{sun2019adversarial}
Bo Sun, Nian-hsuan Tsai, Fangchen Liu, Ronald Yu, and Hao Su.
\newblock Adversarial defense by stratified convolutional sparse coding.
\newblock In {\em CVPR}, pages 11447--11456, 2019.

\bibitem{szegedy2013intriguing}
Christian Szegedy, Wojciech Zaremba, Ilya Sutskever, Joan Bruna, Dumitru Erhan,
  Ian Goodfellow, and Rob Fergus.
\newblock Intriguing properties of neural networks.
\newblock In {\em ICLR}, 2014.

\bibitem{uesato2019labels}
Jonathan Uesato, Jean-Baptiste Alayrac, Po-Sen Huang, Robert Stanforth,
  Alhussein Fawzi, and Pushmeet Kohli.
\newblock Are labels required for improving adversarial robustness?
\newblock {\em NeurIPS}, 2019.

\bibitem{uesato2018adversarial}
Jonathan Uesato, Brendan O’donoghue, Pushmeet Kohli, and Aaron Oord.
\newblock Adversarial risk and the dangers of evaluating against weak attacks.
\newblock In {\em International Conference on Machine Learning}, pages
  5025--5034. PMLR, 2018.

\bibitem{Wang2019}
Jianyu Wang and Haichao Zhang.
\newblock Bilateral adversarial training: Towards fast training of more robust
  models against adversarial attacks.
\newblock In {\em ICCV}, pages 6629--6638, 2019.

\bibitem{wang2019convergence}
Yisen Wang, Xingjun Ma, James Bailey, Jinfeng Yi, Bowen Zhou, and Quanquan Gu.
\newblock On the convergence and robustness of adversarial training.
\newblock In {\em ICML}, 2019.

\bibitem{MART_2020}
Yisen Wang, Difan Zou, Jinfeng Yi, James Bailey, Xingjun Ma, and Quanquan Gu.
\newblock Improving adversarial robustness requires revisiting misclassified
  examples.
\newblock In {\em ICLR}, 2019.

\bibitem{xie2020adversarial}
Cihang Xie, Mingxing Tan, Boqing Gong, Jiang Wang, Alan~L Yuille, and Quoc~V
  Le.
\newblock Adversarial examples improve image recognition.
\newblock In {\em Proceedings of the IEEE/CVF Conference on Computer Vision and
  Pattern Recognition}, pages 819--828, 2020.

\bibitem{xie2019feature}
Cihang Xie, Yuxin Wu, Laurens van~der Maaten, Alan~L Yuille, and Kaiming He.
\newblock Feature denoising for improving adversarial robustness.
\newblock In {\em CVPR}, pages 501--509, 2019.

\bibitem{xu2017feature}
Weilin Xu, David Evans, and Yanjun Qi.
\newblock Feature squeezing: Detecting adversarial examples in deep neural
  networks.
\newblock {\em arXiv:1704.01155}, 2017.

\bibitem{yuan2020ensemble}
Jianhe Yuan and Zhihai He.
\newblock Ensemble generative cleaning with feedback loops for defending
  adversarial attacks.
\newblock In {\em CVPR}, pages 581--590, 2020.

\bibitem{ZagoruykoK16}
Sergey Zagoruyko and Nikos Komodakis.
\newblock Wide residual networks.
\newblock In {\em BMVC}, pages 87.1--87.12, 2016.

\bibitem{zhang2019defense}
Haichao Zhang and Jianyu Wang.
\newblock Defense against adversarial attacks using feature scattering-based
  adversarial training.
\newblock In {\em NeurIPS}, pages 1831--1841, 2019.

\bibitem{TRADES_2019}
Hongyang Zhang, Yaodong Yu, Jiantao Jiao, Eric Xing, Laurent El~Ghaoui, and
  Michael Jordan.
\newblock Theoretically principled trade-off between robustness and accuracy.
\newblock In {\em ICML}, pages 7472--7482, 2019.

\bibitem{zhang2019theoretically}
Hongyang Zhang, Yaodong Yu, Jiantao Jiao, Eric~P. Xing, Laurent~El Ghaoui, and
  Michael~I. Jordan.
\newblock Theoretically principled trade-off between robustness and accuracy.
\newblock In {\em ICML}, 2019.

\bibitem{Zhong2019}
Yaoyao Zhong and Weihong Deng.
\newblock Adversarial learning with margin-based triplet embedding
  regularization.
\newblock In {\em ICCV}, pages 6549--6558, 2019.

\bibitem{zhou2016learning}
Bolei Zhou et~al.
\newblock Learning deep features for discriminative localization.
\newblock In {\em CVPR}, 2016.

\end{thebibliography}
}
\end{document}